\documentclass[11pt]{article}

\usepackage[final]{acl}

\usepackage{times}
\usepackage{latexsym}
\usepackage[T1]{fontenc}
\usepackage[utf8]{inputenc}
\usepackage{microtype}
\usepackage{inconsolata}
\usepackage{graphicx}

\usepackage{subcaption}
\usepackage{booktabs}
\usepackage{amsmath}
\usepackage{amssymb}
\usepackage{mathtools}
\usepackage{amsthm}
\usepackage{array}
\usepackage{multirow}
\usepackage{colortbl}
\usepackage{adjustbox}
\usepackage{makecell}
\usepackage{wrapfig}
\usepackage{algorithm}
\usepackage{algorithmic}
\providecolor{teal}{RGB}{0,128,128}

\newtheorem{assumption}{Assumption}
\newtheorem{proposition}{Proposition}

\usepackage{tabularx}

\usepackage[most]{tcolorbox}
\tcbuselibrary{listings,skins,breakable}
\usepackage{xcolor}

\usepackage[most]{tcolorbox}
\usepackage{fvextra}

\usepackage{booktabs}
\usepackage{multirow}
\usepackage{adjustbox}
\usepackage[table]{xcolor}
\usepackage{makecell}

\DeclareUnicodeCharacter{00A0}{~}
\DeclareUnicodeCharacter{2013}{--}
\DeclareUnicodeCharacter{2014}{---}
\DeclareUnicodeCharacter{2018}{`}
\DeclareUnicodeCharacter{2019}{'}
\DeclareUnicodeCharacter{201C}{``}
\DeclareUnicodeCharacter{201D}{''}
\DeclareUnicodeCharacter{2212}{-}
\DeclareUnicodeCharacter{00E1}{\'a}
\DeclareUnicodeCharacter{00E9}{\'e}
\DeclareUnicodeCharacter{00ED}{\'i}
\DeclareUnicodeCharacter{00C7}{\c{C}}

\title{\textsc{GLIDE}: Generalized Layer-wise Intrinsic Distributional Evaluation for Heterogeneous LLM Agents}

\author{
    Wei Zhu$^{1,2}$, Yiming Wang$^{3}$, Rui Wang$^{3}$, Lixing Yu$^{1,2}$, Kun Yue$^{1,2}$, Zhiwen Tang$^{1,2}$\thanks{Corresponding author.} \\
    $^1$School of Information Science and Engineering, Yunnan University, Kunming, China \\
    $^2$Yunnan Key Laboratory of Intelligent Systems and Computing, Kunming, China \\
    $^3$School of Computer Science, Shanghai Jiao Tong University, Shanghai, China \\
    \texttt{zhuwei@stu.ynu.edu.cn, zhiwen.tang@ynu.edu.cn}
}

\begin{document}
\maketitle

\begin{abstract}
LLM agents require reliable step-level evaluation to compare candidate branches and allocate computation effectively. 
However, lightweight evaluation remains challenging. External verifiers introduce additional inference cost, while agent-produced confidence or self-evaluation scores can be miscalibrated, especially when candidates are generated by heterogeneous agents. 
We propose \textbf{G}eneralized \textbf{L}ayer-wise \textbf{I}ntrinsic \textbf{D}istributional \textbf{E}valuation (\textbf{GLIDE\footnote{Code is available at \url{https://github.com/ZHUWEI-hub/GLIDE}.}}) for LLM agents. 
\textsc{GLIDE} derives intrinsic step evidence from layer-wise residual coherence, which measures whether local residual updates consistently support the global residual change induced by a candidate step. 
It calibrates this evidence against the recent score distribution of the generating agent and converts it into a pessimistic reward that jointly accounts for absolute residual evidence and agent-relative standing. 
The reward provides a cross-agent value signal for MCTS branch selection, while normalized predictive uncertainty guides adaptive branching. 
Experiments on multi-hop reasoning, sequential decision making, and symbolic logic show that \textsc{GLIDE} improves task performance, step-level ranking quality, and computational efficiency without external verifiers or task-specific supervision.
\end{abstract}

\section{Introduction}

Large language models are increasingly used as agents for complex reasoning and interactive decision-making, where success depends on a sequence of intermediate decisions rather than a single generation. 
A locally plausible yet unhelpful step can redirect the agent toward an erroneous trajectory and cause errors to accumulate. 
Recent frameworks mitigate this brittleness by exploring multiple candidate continuations through tree-structured reasoning, MCTS-style planning, and multi-agent trajectory exploration~\citep{tot,hao2023reasoning,zhou2024language,symphony}. 
Yet exploration is only useful when candidate branches can be reliably compared and computation can be allocated to promising directions. 
Inaccurate, sparse, or poorly calibrated comparison signals may repeatedly favor misleading trajectories, making intermediate-step evaluation a central bottleneck in inference-time agent control.

\begin{figure}[t!] 
    \centering
    \includegraphics[width=\columnwidth]{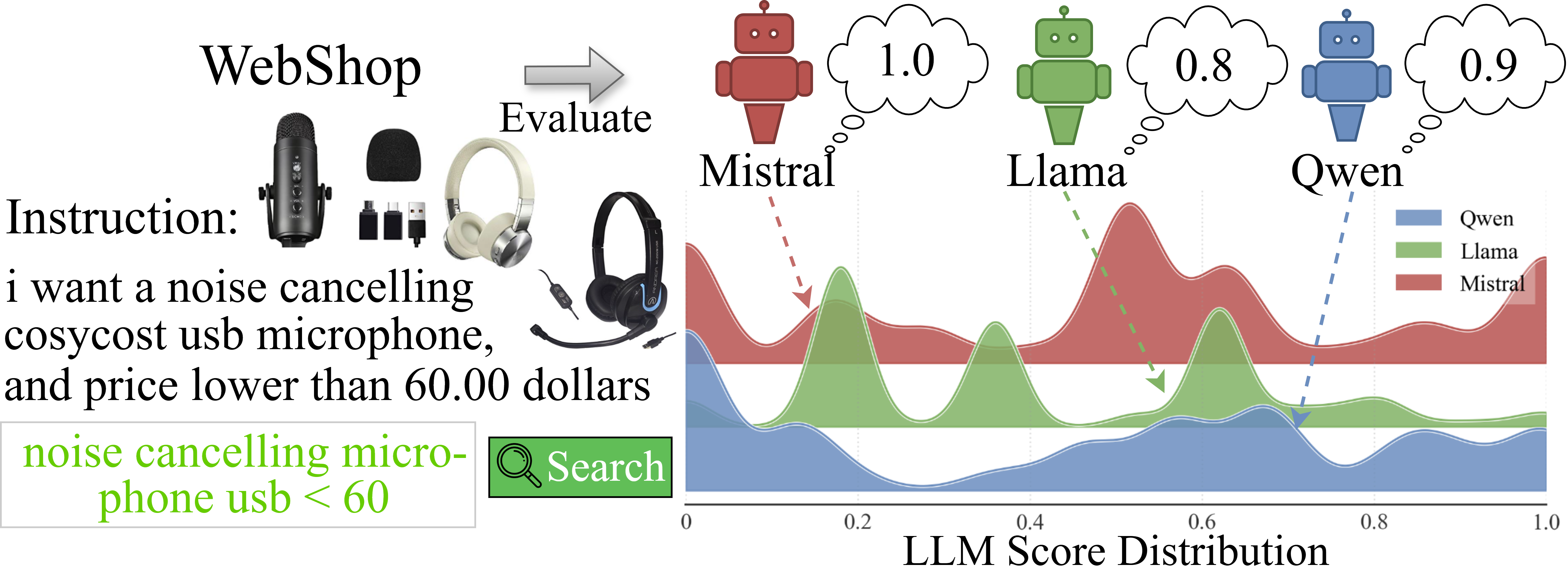} 
\caption{
\textbf{Score calibration mismatch across heterogeneous agents.}
Self-generated score distributions show that different LLM agents produce incompatible confidence scales on comparable inputs, making raw scores unreliable for cross-agent comparison.
}
    \label{fig:motivation}

\end{figure}

A natural solution is to train Process Reward Models, which provide dense step-level supervision through dedicated discriminators~\citep{skywork,zhang2024rest,zhao2025genprm}.
While effective in controlled settings, PRMs require task-specific supervision data, additional training procedures, and extra verifier inference.
These requirements make them difficult to deploy as lightweight, general-purpose evaluators for online branch comparison.
As a result, many inference-time systems turn to lighter alternatives based on prompt-based self-evaluation or LLM-as-a-Judge~\citep{gu2024survey,zhuge2025agent}, where an LLM is asked to score or compare candidate steps without training a separate reward model.

However, judge-based evaluation introduces its own limitations.
Its judgments are bounded by the reasoning capability and calibration behavior of the evaluator model, and it may assign high scores to fluent yet incorrect intermediate steps.
The issue becomes more pronounced in heterogeneous agent pools~\citep{park2025ensembling,ye2025x,symphony}, where candidate branches may be generated by agents built on different base models, prompting policies, or decoding behaviors.
Different agents may contribute complementary reasoning patterns, action preferences, and failure modes, but they also produce confidence estimates, self-evaluation scores, or other generator-dependent signals on incompatible scales.
As illustrated in Figure~\ref{fig:motivation}, a numerically high score from one agent may reflect calibration bias rather than a genuinely better candidate step.
Consequently, lightweight evaluation in heterogeneous agent inference faces a practical dilemma.
External evaluators can provide a more shared scoring basis but are costly to invoke at every candidate step, whereas generator-dependent signals are inexpensive but cannot be directly compared across agents.

This dilemma motivates an evaluation signal that is intrinsic to generation, inexpensive to compute, and calibrated before being used for cross-agent comparison.
Rather than invoking an external verifier, we examine whether useful candidate steps leave consistent signatures in the generator's internal computation.
Under the residual-stream view of Transformers~\citep{elhage2021mathematical}, layer-wise updates write additive changes into a shared representation stream.
We find that useful candidate steps tend to exhibit stable local-to-global residual coherence, meaning that layer-wise updates consistently align with the global update induced by the candidate step.
Failure-specific transitions, in contrast, show weaker or less stable agreement between local layer updates and the overall step update.
This regularity suggests an intrinsic indicator of step-level usefulness that can be computed from the generator's forward pass without task-specific supervision or additional judge calls.

Building on this observation, we propose \textsc{GLIDE}, \textbf{G}eneralized \textbf{L}ayer-wise \textbf{I}ntrinsic \textbf{D}istributional \textbf{E}valuation for heterogeneous LLM-agent MCTS.
\textsc{GLIDE} first derives a step-level intrinsic preference signal from layer-wise residual coherence, capturing whether local hidden-state updates consistently support the global update induced by a candidate step.
It then calibrates this signal against each agent's recent score distribution and converts it into a pessimistic reward, making candidate steps produced by different agents more reliably comparable.
The calibrated reward is used for MCTS value estimation and further combined with predictive uncertainty to support adaptive branching.
As a result, \textsc{GLIDE} enables intrinsic node evaluation and resource allocation without external verifiers, task-specific supervision, or additional judge passes.

Our contributions are summarized as follows:

\begin{itemize}
    \item We identify layer-wise residual coherence as an intrinsic regularity for step-level preference estimation.
    \item We propose \textsc{GLIDE}, which combines intrinsic layer-wise evidence with agent-wise distributional calibration for heterogeneous LLM-agent MCTS.
    \item Experiments show that \textsc{GLIDE} improves performance-cost trade-offs and step-level preference quality.
\end{itemize}

\section{Related Work}

\paragraph{Agentic Reasoning and Planning}
LLM reasoning has shifted from linear generation to structured search and planning, including CoT-style rationale elicitation~\citep{wei2022chaincot}, interactive refinement with ReAct~\citep{yao2023react} and Reflexion~\citep{shinn2023reflexion}, and multi-path aggregation~\citep{wang2022self-con,zhang2023end}. 
Recent work further formulates reasoning as state-space search, covering tree/graph exploration such as ToT~\citep{tot}, GoT~\citep{got}, and MDToC~\citep{ta2025mdtoc}, MCTS-style planners such as RAP~\citep{hao2023reasoning} and LATS~\citep{zhou2024language}, and heterogeneous multi-agent systems such as SYMPHONY~\citep{symphony}, MoA~\citep{wang2024mixture_moa}, and MASTER~\citep{gan2025master}. 
However, these systems still rely on reliable value estimation, where judge-based scoring adds inference cost and heterogeneous agents introduce non-comparable score scales. 
This motivates intrinsic, calibrated search-control signals without external verifiers.


\paragraph{Verification Mechanisms and Process Supervision}

Intermediate-step evaluation mainly follows two lines: extrinsic process supervision and intrinsic model signals. 
PRM-based methods provide dense step-level feedback through trained discriminators~\citep{wang2024math,skywork}, but even with automated synthesis or self-training~\citep{zhang2025groundedprm,zhang2024rest}, they require task-specific data, extra training, and additional inference. 
Intrinsic alternatives based on token-level statistics, including perplexity, entropy, and energy scores~\citep{si2022prompting,huang2023look,malinin2020uncertainty}, avoid these costs but largely reflect local confidence and can fail under fluent hallucinations. 
Recent hidden-state methods support trace diagnosis or trajectory pruning~\citep{trace,clue,step,cotkinetics,2026-icml}, and CoE~\citep{coe} studies hidden-state trajectories, but they mainly target full-chain analysis or local smoothness rather than online step-wise ranking for calibrated heterogeneous control. 
This motivates a supervision-free intrinsic signal tailored to step-wise heterogeneous search.

\begin{figure*}[t]
    \centering
    \includegraphics[width=0.95\textwidth]{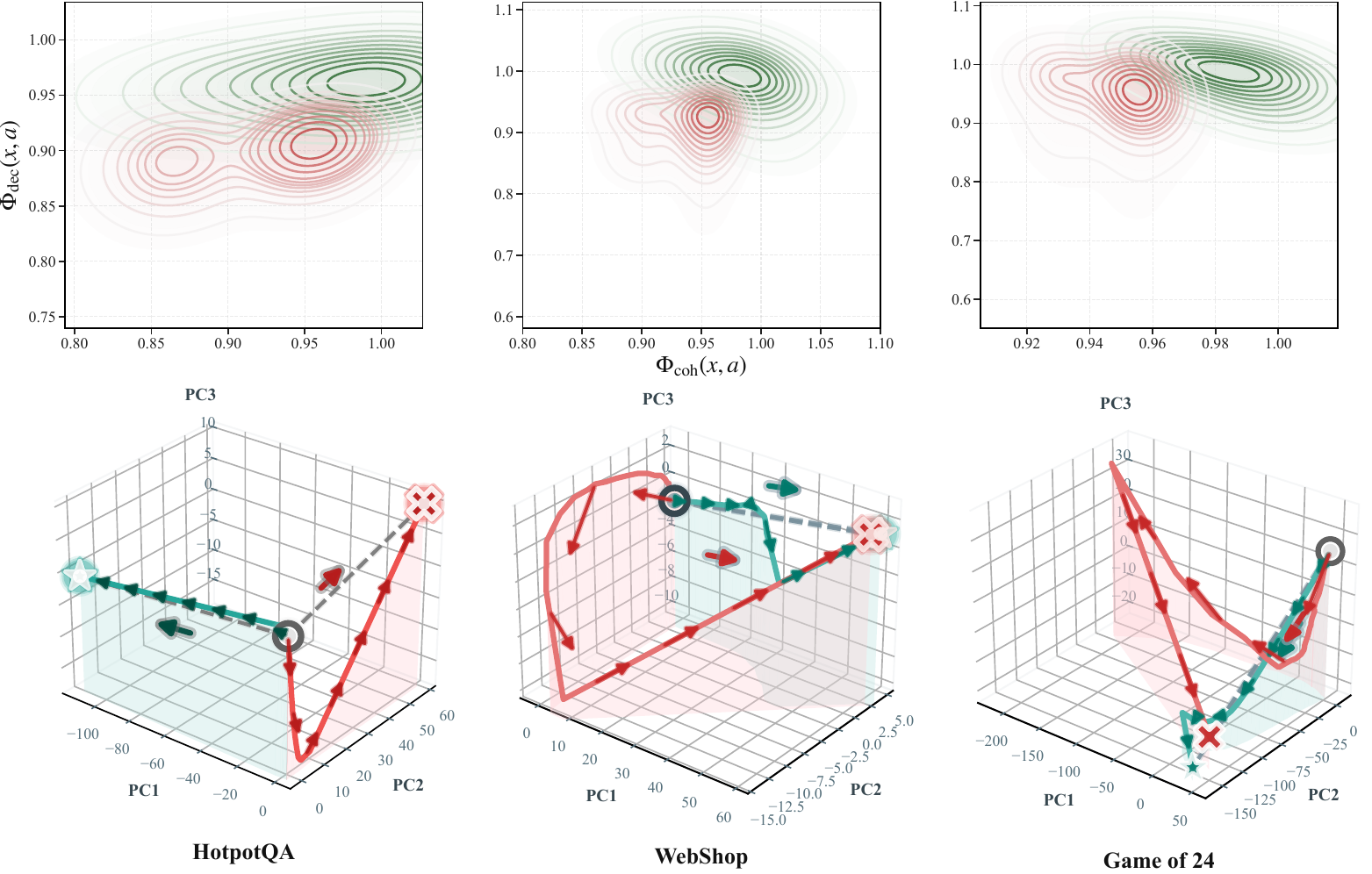}
\caption{
\textbf{Diagnostics of intrinsic stability in layer-wise evidence.}
\textbf{Top}: Transitions from successful trajectories (Green) concentrate in high-coherence/high-decisiveness regions, while transitions unique to failed trajectories (Red) are dispersed in the $(\Phi_{\mathrm{coh}}, \Phi_{\mathrm{dec}})$ space. 
\textbf{Bottom}: Successful trajectories (Green) exhibit stable layer-wise residual evidence, whereas failed trajectories (Red) fluctuate and do not consistently progress toward a shared direction.
}
    \label{fig:irc_diagnostics}
\end{figure*}

\section{GLIDE: Generalized Layer-wise Intrinsic Distributional Evaluation}

\begin{figure*}[t]
  \centering  
  \includegraphics[width=\linewidth]{img/EMNLP-overview.drawio.png}  

  \caption{
    Overview of GLIDE
  }
  \label{fig:overview}  
\end{figure*}

\textsc{GLIDE} constructs a calibrated step-level reward for heterogeneous LLM-agent search without reward-model training or external judge calls. 
Specifically, it measures layer-wise residual coherence as the intrinsic stability score $S_{\mathrm{ISLE}}$, normalizes this score within the generating agent's recent behavior, and forms a pessimistic reward $R_{\mathrm{GLIDE}}$ for cross-agent value comparison. 
The calibrated reward drives MCTS selection, and agent-relative predictive uncertainty further determines how many successors should be expanded. 
The full algorithm is given in Appendix~\ref{appendix:algorithm}.




\subsection{Layer-wise Intrinsic Stability Evaluation}
\label{subsec:residual_motivation}
\label{subsec:irc_score_theory}
\label{subsec:empirical_rc}

We construct an intrinsic step-level signal from the generator's forward computation by measuring whether a candidate step is supported by stable residual evidence. 
Under the residual-stream view of Transformers~\citep{elhage2021mathematical}, each layer writes an additive update into a shared representation stream, and prior work suggests that hidden-state dynamics contain structured signals related to reasoning behavior~\citep{coe,wang2025sampling,zhu2025survey,geva2022transformer,dar2023analyzing}. 
Thus, a useful step should not induce layer-wise updates that behave as unrelated perturbations; instead, these updates should coherently accumulate toward the global residual change associated with the step.

Consider an input state $x$ and a candidate step $a$ realized as a continuation $y_{1:\ell}$.
All quantities are conditioned on the transition $(x,a)$.
Let $\mathbf{H}^{(l)}\in\mathbb{R}^{\ell\times d}$ denote the hidden states of the generated continuation at layer $l$.
We summarize the layer representation of this step by
$
\mathbf{h}^{(l)}
=
\frac{1}{\ell}
\sum_{j=1}^{\ell}
\mathbf{H}_{j}^{(l)}
\label{eq:layer_pooling}
$. 
The local residual update at layer $l$ and the global residual update are defined as
\begin{equation}
\mathbf{u}_{l}
=
\mathbf{h}^{(l+1)}-\mathbf{h}^{(l)},
\quad
\mathbf{u}_{\mathrm{global}}
=
\mathbf{h}^{(L)}-\mathbf{h}^{(0)}.
\label{eq:residual_updates}
\end{equation}


We quantify layer-wise support through the directional agreement between each local update 
$\mathbf{u}_{l}$ and the global residual update 
$\mathbf{u}_{\mathrm{global}}$.
This choice is meaningful when the residual updates of a candidate step are not arbitrary perturbations, but share a coherent low-dimensional structure and accumulate toward the same step-level change.
We state this condition as follows.

\begin{assumption}[Coherent residual accumulation]
\label{assump:coherent_residual_accumulation}
For a fixed candidate step, the layer-wise residual updates 
$\{\mathbf{u}_l\}_{l=0}^{L-1}$ concentrate near a shared low-dimensional residual subspace, within which a dominant component accumulates consistently across layers with non-vanishing magnitude.
\end{assumption}

Under this assumption, a candidate step is supported by residual evidence that is both locally structured and progressively accumulated.
The concentration condition prevents the layer-wise updates from behaving as independent high-dimensional fluctuations, while the dominant-component condition ensures that their contributions do not vanish or cancel across layers.
Consequently, the average local-to-global agreement provides a natural proxy for the stability of intrinsic step evidence.
Appendix~\ref{app:residual_subspace} and Appendix~\ref{app:coherence_margin} provide the corresponding low-rank derivation and margin analysis.
We therefore define the residual-coherence term as

{\small
\begin{equation}
\Phi_{\mathrm{coh}}(x,a)
=
\operatorname{sigmoid}
\left(
\frac{\sqrt{d}}{L}
\sum_{l=0}^{L-1}
\cos(\mathbf{u}_l,\mathbf{u}_{\mathrm{global}})
\right),
\label{eq:coherence_score}
\end{equation}
}
where the cosine terms measure local-to-global directional support, and the layer average captures whether such support is persistent across the computation.
The factor $\sqrt{d}$ compensates for the concentration of high-dimensional cosine values~\citep{vershynin2018high}, while the sigmoid maps the aggregated evidence to a bounded scale.

Residual coherence alone may still assign a high value to an internally consistent but weakly determined continuation.
We therefore introduce token-level decisiveness as a conservative gate rather than as an independent evaluator:
{\small
\begin{equation}
\Phi_{\mathrm{dec}}(x,a)
=
\left(
\frac{1}{\ell}
\sum_{j=1}^{\ell}
\max_{w\in\mathcal{V}}
p_{\theta}(w\mid x,y_{<j})
\right)^2 .
\label{eq:decisiveness_score}
\end{equation}
}
This term downweights steps generated under diffuse predictive distributions.
The square makes the gate stricter, so that a candidate obtains a high score only when the model is both internally coherent and locally decisive.

We then define the \textbf{I}ntrinsic \textbf{S}tability of \textbf{L}ayer-wise \textbf{E}vidence (ISLE)  score, denoted as $S_{\mathrm{ISLE}}$, by
\begin{equation}
S_{\mathrm{ISLE}}(x,a)
=
\Phi_{\mathrm{coh}}(x,a)
\cdot
\Phi_{\mathrm{dec}}(x,a). 
\label{eq:isle_score}
\end{equation}

A high $S_{\mathrm{ISLE}}$ therefore requires two forms of evidence to agree. Layer-wise residual updates should consistently support the global residual update, and the decoder should produce the continuation with sufficient local decisiveness.
Because both terms are computed from the same forward pass, ISLE does not require task-specific supervision or additional verifier calls.

As a sanity check, we examine whether the intrinsic stability evidence terms are associated with successful transitions in the inference traces.
Figure~\ref{fig:irc_diagnostics} shows that successful transitions are more concentrated in regions with high $\Phi_{\mathrm{coh}}$ and high $\Phi_{\mathrm{dec}}$, whereas transitions that appear only in failed trajectories are more diffuse and exhibit weaker residual evidence.
This diagnostic provides empirical support for the coherent residual accumulation assumption behind $S_{\mathrm{ISLE}}$.


\subsection{Distributional Calibration for Heterogeneous Agents}
\label{subsec:calibration}

The intrinsic score $S_{\mathrm{ISLE}}$ provides an agent-local step signal, but heterogeneous search requires comparing candidates generated by different agents. 
Let $\mathcal{M}=\{M_i\}_{i=1}^{B}$ be the agent pool, and denote the score of a step $a$ generated by $M_i$ at state $x$ as $S_{\mathrm{ISLE}}^{(i)}(x,a)$. 
Within a fixed agent, this score is expected to preserve local preference information because it measures whether the agent's residual evidence coherently supports the generated transition.

\begin{proposition}[Agent-local preference consistency of $S_{\mathrm{ISLE}}$]
\label{prop:preference_consistency}
Consider two candidate steps $a$ and $a'$ generated by the same agent $M_i$ at state $x$.
Let $q(x,a)$ denote the latent transition utility of $a$, and define
$
\Delta_q(a,a') = q(x,a)-q(x,a'),
\Delta_S^{(i)}(a,a')
=
S_{\mathrm{ISLE}}^{(i)}(x,a)
-
S_{\mathrm{ISLE}}^{(i)}(x,a').
$

Assume that, within agent $M_i$, $S_{\mathrm{ISLE}}^{(i)}$ is an order-preserving noisy surrogate of $q$ with sub-Gaussian perturbation.
Then whenever $\Delta_q(a,a')>0$,  there exists an agent-dependent constant $\kappa_i>0$ such that, 
{\small
\begin{equation}
\Pr\!\left(
\Delta_S^{(i)}(a,a') \le 0
\right)
\le
\exp\!\left(
-\kappa_i \Delta_q(a,a')^2
\right), 
\label{eq:preference_consistency_bound}
\end{equation}
}
\end{proposition}

Proposition~\ref{prop:preference_consistency} states that the probability of a ranking reversal, $\Pr(\Delta_S^{(i)}(a,a')\le 0)$, decreases exponentially with the latent utility gap $\Delta_q(a,a')$.
The full proof is provided in Appendix~\ref{app:preference_consistency}.

However, agent-local consistency does not ensure cross-agent comparability, since different agents may exhibit different hidden-state dynamics, confidence scales, and intrinsic score distributions. 
Thus, heterogeneous candidate selection must retain the local preference information of $S_{\mathrm{ISLE}}$ while correcting agent-specific scale mismatch.

We reduce this mismatch by rank-normalizing scores within each agent. 
For each agent $M_i$, we maintain a FIFO window $\mathcal{W}_i^{S}$ of recent $S_{\mathrm{ISLE}}$  produced by that agent and define the empirical rank 
{\small
\begin{equation}
S_{\mathrm{rank}}^{(i)}(x,a)
=
\frac{1}{|\mathcal{W}_i^{S}|}
\sum_{\chi \in \mathcal{W}_i^{S}}
\mathbb{I}
\!\left[
\chi
\le
S_{\mathrm{ISLE}}^{(i)}(x,a)
\right],
\label{eq:rank_score}
\end{equation}
}
which measures its standing within the generating agent's recent score distribution.

Rank normalization improves cross-agent comparability, but using rank alone would discard absolute intrinsic strength.
A step may rank highly only because its agent has recently produced weak candidates, while another step may have strong residual evidence but appear less exceptional under a more confident generator.
We therefore combine the raw score, which reflects absolute residual evidence, with the rank score, which reflects agent-relative standing. 
The \textsc{GLIDE} reward is defined by harmonic fusion:
{\small
\begin{equation}
R_{\mathrm{GLIDE}}^{(i)}(x,a)
=
\frac{
2S_{\mathrm{ISLE}}^{(i)}(x,a)
S_{\mathrm{rank}}^{(i)}(x,a)
}{
S_{\mathrm{ISLE}}^{(i)}(x,a)
+
S_{\mathrm{rank}}^{(i)}(x,a)
}.
\label{eq:glide_reward}
\end{equation}
}

Because harmonic fusion is dominated by the smaller term, $R_{\mathrm{GLIDE}}^{(i)}$ is pessimistic: a step receives a high reward only when it has both strong intrinsic evidence and high agent-relative standing. 
This converts the agent-local signal justified by Proposition~\ref{prop:preference_consistency} into a calibrated cross-agent comparison signal for value estimation and adaptive branching.



\begin{table*}[t]  
\centering  
\small      

\begin{minipage}[t]{0.31\linewidth}
\centering  
\caption[HotpotQA]{HotpotQA (Hot).}
\label{tab:hotpotqa}
\setlength{\tabcolsep}{2pt}  
\begin{tabular}{@{}lc@{}}
\toprule
\textbf{Method} & \textbf{Exact Match $\uparrow$} \\
\midrule
CoT             & 0.34 \\
CoT-SC         & 0.38 \\
ReAct         & 0.39 \\
Reflexion  & 0.51 \\
ToT             & 0.55 \\
RAP       & 0.60 \\
LATS         & 0.71 \\
Beam Retrieval   & 0.73 \\
MASTER         & 0.76 \\
SYMPHONY-S & 0.59 \\
SYMPHONY-L & 0.79 \\
\midrule
\textbf{GLIDE} & \textbf{0.73} \\
\bottomrule
\end{tabular}
\end{minipage}
\hspace{0.01\linewidth}  
\begin{minipage}[t]{0.31\linewidth}
\centering
\caption[WebShop]{WebShop (Web).}
\label{tab:webshop_results}
\setlength{\tabcolsep}{2pt}
\begin{tabular}{lcc}
\toprule
\textbf{Method} & \textbf{Score $\uparrow$} & \textbf{SR $\uparrow$} \\
\midrule
IL           & 0.60 & 0.29 \\
IL+RL       & 0.62 & 0.29 \\
ReAct      & 0.54 & 0.32 \\
Reflexion  & 0.64 & 0.35 \\
MoA   & 0.31 & 0.42 \\
WebGUM   & 0.68 & 0.45 \\
AgentKit       & 0.70 & --  \\
LATS       & 0.76 & 0.38 \\
MASTER           & 0.80 & -- \\
Human Expert    & 0.82 & 0.60 \\
SYMPHONY-S           & 0.82 & 0.56 \\
SYMPHONY-L           & 0.88 & 0.72 \\

\midrule
\textbf{GLIDE} & \textbf{0.84} & \textbf{0.57}\\

\bottomrule
\end{tabular}
\end{minipage}
\hspace{0.01\linewidth}  
\begin{minipage}[t]{0.31\linewidth}
\centering
\caption[Game of 24]{Game of 24 (G24).}
\label{tab:mbpp_results}
\setlength{\tabcolsep}{2pt}
\begin{tabular}{lc}  
\toprule
\textbf{Method} & \textbf{Success Rate $\uparrow$} \\
\midrule
Llama-2(70B)  & 0.04 \\
Gemini pro    & 0.08 \\
Claude-3 Opus    & 0.07 \\
GPT-4 turbo     & 0.09 \\
CoT   & 0.08 \\
Reflexion   & 0.12 \\
ToT   & 0.20 \\
GoT    & 0.21 \\
MoA   & 0.13 \\
RAP     & 0.40 \\
MDToC    & 0.30 \\
LATS     & 0.44 \\
SYMPHONY-S           & 0.43 \\

\midrule
\textbf{GLIDE} & \textbf{0.54} \\

\bottomrule
\end{tabular}
\end{minipage}

\vspace{5pt}
\centering
\small
Note: Metrics are normalized to the [0,1] range; A dash (–) marks those not reported in the publication.

\end{table*}

\subsection{Adaptive Branching with Calibrated Rewards}
\label{subsec:adaptive_branching}

Given the calibrated reward $R_{\mathrm{GLIDE}}$, \textsc{GLIDE} uses it as the transition value in MCTS and further derives an adaptive branching rule for computation allocation. 
For a candidate step $a$ generated by agent $M_i$ at state $x$, $R_{\mathrm{GLIDE}}^{(i)}(x,a)$ initializes the value of the expanded transition, which is then backpropagated through Eq.~\eqref{eq:back} and used by the UCT objective in Eq.~\eqref{eq:uct}. 
Thus, branch selection is driven by a verifier-free value signal that is intrinsic to the generator and calibrated within the corresponding agent distribution.

To support online calibration, \textsc{GLIDE} maintains two agent-wise FIFO windows: $\mathcal{W}_i^{S}$ for recent $S_{\mathrm{ISLE}}$ values and $\mathcal{W}_i^{H}$ for recent predictive entropies. 
The former is used for empirical rank calibration in Eq.~\eqref{eq:rank_score}, while the latter provides an agent-relative uncertainty scale. 
New observations are inserted only after the current reward and uncertainty are computed, avoiding self-normalization.

While $R_{\mathrm{GLIDE}}$ estimates the exploitation value of a transition, expansion width should also depend on whether its successor still requires exploration. 
We therefore allocate computation by coupling agent-relative predictive uncertainty with reward saturation. 
For a continuation $y_{1:\ell}$ generated by $M_i$, let $p_i$ denote its token distribution and compute
$
\bar{\mathcal{H}}_i(x,a)
=
-\frac{1}{\ell}
\sum_{j=1}^{\ell}
\sum_{w\in\mathcal{V}}
p_i(w\mid x,y_{<j})
\log p_i(w\mid x,y_{<j}) .
$
Given the mean $\mu_i^{H}$ and standard deviation $\sigma_i^{H}$ of $\mathcal{W}_i^{H}$, the normalized uncertainty is
{\small
\begin{equation}
\widetilde{\mathcal{H}}_i(x,a)
=
\operatorname{sigmoid}
\left(
(\bar{\mathcal{H}}_i(x,a)-\mu_i^{H})/\sigma_i^{H}
\right).
\label{eq:relative_entropy}
\end{equation}
}
The branching width for the successor induced by $(x,a)$ is then
\begin{equation}
\begin{aligned}
n(x,a)
=
\Big\lfloor
&n_{\min}
+
(n_{\max}-n_{\min})
\widetilde{\mathcal{H}}_i(x,a)
\\
&\cdot
\left[
1-
\left(R_{\mathrm{GLIDE}}^{(i)}(x,a)\right)^2
\right]
\Big\rfloor .
\end{aligned}
\label{eq:adaptive_branching}
\end{equation}


This formulation favors uncertain, insufficiently supported successors while suppressing redundant expansion near high-reward regions. 
Given $n(x,a)$, \textsc{GLIDE} uniformly samples generators from the heterogeneous pool to preserve diversity without a learned router.

Overall, \textsc{GLIDE} provides a calibrated intrinsic reward that makes generator-internal step evidence usable for heterogeneous MCTS control. 
The complete Algorithm is given in Appendix~\ref{appendix:algorithm}.

\begin{table*}[t]
    \centering
    \footnotesize
    \setlength{\tabcolsep}{3.5pt}
    
    \definecolor{AvgGray}{gray}{0.92}
    \definecolor{LineGray}{gray}{0.7}

\caption{
\textbf{Step-level preference quality of $R_{\mathrm{GLIDE}}$.}
We report ranking metrics (P@1, MRR, NDCG@3) on expanded child nodes from HotpotQA (Hot), WebShop (Web), and Game of 24 (G24), with Avg denoting the mean.
Efficiency reports per-candidate scoring overhead; \textbf{\checkmark} indicates negligible overhead.
Hetero-Judge randomly selects scorers from the heterogeneous agent pool.
Baselines are detailed in Appendix~\ref{app:discriminative_metrics}.
}
    \label{tab:discriminative_metrics_full}

    \begin{tabular}{ll|cccc|cccc|cccc||cc}
        \toprule
        \multirow{2}{*}{\textbf{Category}} & \multirow{2}{*}{\textbf{Method}} & \multicolumn{4}{c|}{\textbf{P@1}} & \multicolumn{4}{c|}{\textbf{MRR}} & \multicolumn{4}{c||}{\textbf{NDCG@3}} & \multicolumn{2}{c}{\textbf{Efficiency}} \\
        
        \cmidrule(lr){3-6} \cmidrule(lr){7-10} \cmidrule(lr){11-14} \cmidrule(l){15-16}
        
        & & Hot & Web & G24 & \cellcolor{AvgGray}\textbf{Avg} & Hot & Web & G24 & \cellcolor{AvgGray}\textbf{Avg} & Hot & Web & G24 & \cellcolor{AvgGray}\textbf{Avg} & \textbf{Time} & \textbf{Mem} \\
        \midrule
        
        \multirow{5}{*}{\textit{Intrinsic}} 
         & Perplexity & 0.17 & 0.37 & 0.38 & \cellcolor{AvgGray}0.31 & 0.52 & 0.66 & 0.64 & \cellcolor{AvgGray}0.61 & 0.59 & 0.71 & 0.65 & \cellcolor{AvgGray}0.65 & \checkmark & \checkmark \\
         & Energy & 0.24 & 0.00 & 0.00 & \cellcolor{AvgGray}0.08 & 0.56 & 0.44 & 0.33 & \cellcolor{AvgGray}0.44 & 0.63 & 0.30 & 0.30 & \cellcolor{AvgGray}0.41 & \checkmark & \checkmark \\
         & LN-Entropy & 0.33 & 0.00 & 0.00 & \cellcolor{AvgGray}0.11 & 0.61 & 0.44 & 0.33 & \cellcolor{AvgGray}0.46 & 0.69 & 0.30 & 0.30 & \cellcolor{AvgGray}0.43 & \checkmark & \checkmark \\
         & CoE-R & 0.44 & 0.48 & 0.40 & \cellcolor{AvgGray}0.44 & 0.68 & 0.72 & 0.64 & \cellcolor{AvgGray}0.68 & 0.74 & 0.79 & 0.62 & \cellcolor{AvgGray}0.72 & \checkmark & \checkmark \\
         & CoE-C & 0.46 & 0.47 & 0.41 & \cellcolor{AvgGray}0.45 & 0.69 & 0.71 & 0.65 & \cellcolor{AvgGray}0.68 & 0.75 & 0.78 & 0.66 & \cellcolor{AvgGray}0.73 & \checkmark & \checkmark \\
        \midrule

        \multirow{3}{*}{\textit{Extrinsic}} 
         & PRM-RLHFlow & 0.36 & 0.47 & 0.36 & \cellcolor{AvgGray}0.40 & 0.63 & 0.72 & 0.63 & \cellcolor{AvgGray}0.66 & 0.68 & 0.77 & 0.63 & \cellcolor{AvgGray}0.69 & 0.10s & 13.7k \\
         & GenPRM-7B & 0.41 & 0.36 & 0.30 & \cellcolor{AvgGray}0.36 & 0.66 & 0.66 & 0.60 & \cellcolor{AvgGray}0.64 & 0.72 & 0.73 & 0.60 & \cellcolor{AvgGray}0.68 & 0.10s & 14.2k \\
         & PRM-Skywork  & 0.32 & 0.40 & 0.48 & \cellcolor{AvgGray}0.40 & 0.62 & 0.67 & 0.70 & \cellcolor{AvgGray}0.66 & 0.68 & 0.74 & 0.67 & \cellcolor{AvgGray}0.70 & 0.04s & 3.0k \\
        \midrule

        \textit{Multi-Agent} 
         & Hetero-Judge & 0.45 & 0.56 & 0.41 & \cellcolor{AvgGray}0.47 & 0.69 & 0.74 & 0.66 & \cellcolor{AvgGray}0.70 & 0.71 & 0.80 & 0.67 & \cellcolor{AvgGray}0.73 & 4.63s & 15.5k \\
        \midrule
        
        \textit{Ours} 
         & \textbf{GLIDE} & \textbf{0.59} & \textbf{0.48} & \textbf{0.43} & \cellcolor{AvgGray}\textbf{0.50} & \textbf{0.77} & \textbf{0.73} & \textbf{0.67} & \cellcolor{AvgGray}\textbf{0.72} & \textbf{0.83} & \textbf{0.79} & \textbf{0.71} & \cellcolor{AvgGray}\textbf{0.78} & \textbf{\checkmark} & \textbf{\checkmark} \\
        \bottomrule
    \end{tabular}
    
\end{table*}

\begin{table*}[t]
    \centering
    \small
    \setlength{\tabcolsep}{4.2pt}
\caption{
\textbf{Controlled comparison across agent pools.}
Pools use $L$ (Llama-3.1-8B-Instruct), $M$ (Mistral-7B-Instruct-v0.3), and $Q$ (Qwen2.5-7B-Instruct-1M).
Performance uses EM, SR, and Score; Cost denotes total input/output tokens in millions.
\textcolor{teal}{Teal} marks the better result between SYMPHONY (Sym) and \textsc{GLIDE}.
}  
    \label{tab:main_comparison_full}
    
    \newcommand{\win}[1]{\textcolor{teal}{#1}}
    
    \begin{tabular}{l | cc cc | cc cc cc | cc cc}
        \toprule
        & \multicolumn{4}{c|}{\textbf{HotpotQA (Hot)}} 
        & \multicolumn{6}{c|}{\textbf{WebShop (Web)}} 
        & \multicolumn{4}{c}{\textbf{Game of 24 (G24)}} \\
        
        & \multicolumn{2}{c}{EM ($\uparrow$)} & \multicolumn{2}{c|}{Cost (M) ($\downarrow$)} 
        & \multicolumn{2}{c}{SR ($\uparrow$)} & \multicolumn{2}{c}{Score ($\uparrow$)} & \multicolumn{2}{c|}{Cost (M) ($\downarrow$)} 
        & \multicolumn{2}{c}{SR ($\uparrow$)} & \multicolumn{2}{c}{Cost (M) ($\downarrow$)} \\
        
        \cmidrule(lr){2-3} \cmidrule(lr){4-5} 
        \cmidrule(lr){6-7} \cmidrule(lr){8-9} \cmidrule(lr){10-11} 
        \cmidrule(lr){12-13} \cmidrule(lr){14-15}
        
        \textbf{Pool Combo} 
        & Sym & \textbf{Ours} & Sym & \textbf{Ours} 
        & Sym & \textbf{Ours} & Sym & \textbf{Ours} & Sym & \textbf{Ours} 
        & Sym & \textbf{Ours} & Sym & \textbf{Ours} \\
        \midrule
        
        \textit{Single: $\{L\}$} 
        & 0.27 & \win{0.36} & 28.5 & \win{18.7} 
        & 0.08 & \win{0.19} & 0.21 & \win{0.42} & \win{102.0} & 175.7 
        & 0.21 & \win{0.30} & 4.8 & \win{3.1} \\
        
        \textit{Single: $\{M\}$} 
        & 0.32 & \win{0.44} & \win{11.1} & 14.9 
        & 0.35 & 0.35 & 0.65 & \win{0.67} & 52.9 & \win{48.0} 
        & 0.15 & \win{0.23} & 25.5 & \win{11.5} \\
        
        \textit{Single: $\{Q\}$} 
        & 0.30 & \win{0.49} & 53.3 & \win{12.1} 
        & 0.39 & \win{0.46} & 0.66 & \win{0.71} & 54.4 & \win{20.2} 
        & 0.14 & 0.14 & 9.6 & \win{4.6} \\
        
        \arrayrulecolor{black!30}\midrule\arrayrulecolor{black}
        
        \textit{Dual: $\{L,M\}$} 
        & 0.52 & \win{0.53} & 24.4 & \win{12.0} 
        & 0.37 & \win{0.45} & 0.73 & \win{0.77} & 82.5 & \win{44.0} 
        & 0.31 & \win{0.40} & \win{6.4} & 6.5 \\
        
        \textit{Dual: $\{M,Q\}$} 
        & 0.50 & \win{0.62} & 21.5 & \win{10.7} 
        & 0.46 & \win{0.52} & 0.74 & \win{0.80} & 47.6 & \win{29.2} 
        & 0.21 & \win{0.36} & 18.7 & \win{2.5} \\
        
        \textit{Dual: $\{L,Q\}$} 
        & 0.47 & \win{0.60} & 19.9 & \win{13.2} 
        & 0.40 & \win{0.53} & 0.74 & \win{0.81} & 68.7 & \win{25.6} 
        & 0.30 & \win{0.47} & 4.7 & \win{2.8} \\
        
        \arrayrulecolor{black!30}\midrule\arrayrulecolor{black}
        
        \textit{Full: $\{L,M,Q\}$} 
        & 0.59 & \win{0.73} & 18.7 & \win{10.5} 
        & 0.56 & \win{0.57} & 0.82 & \win{0.84} & 139.3 & \win{26.6} 
        & 0.43 & \win{0.54} & 6.3 & \win{4.6} \\
        \bottomrule
    \end{tabular}
\end{table*}

\section{Experiments}
\label{sec:experiments}

\subsection{Experimental Settings}
\label{subsec:settings}

Following SYMPHONY-S~\citep{symphony}, we use a heterogeneous open-source model pool compatible with consumer-grade hardware: Qwen2.5-7B-Instruct-1M~\citep{model:qwen}, Mistral-7B-Instruct-v0.3~\citep{jiang2023mistral7b}, and Llama-3.1-8B-Instruct~\citep{grattafiori2024llama3herdmodels}.
Unless otherwise stated, all controlled task experiments follow prior protocols~\citep{shinn2023reflexion,tot,zhou2024language,gan2025master,symphony}, using the same prompt formats, decoding settings, and search budgets as SYMPHONY-S.
We report the average performance over three runs.
For \textsc{GLIDE}, we set $|\mathcal{W}_i^{S}|=|\mathcal{W}_i^{H}|=50$ and use $n_{\min}=1$, $n_{\max}=4$ in Eq.~\eqref{eq:adaptive_branching}, matching the maximum fixed expansion width of SYMPHONY-S.
Additional implementation details are provided in Appendix~\ref{appendix:main_experiment_details}.

\begin{figure*}[t]
    \centering
    \includegraphics[width=0.99\textwidth]{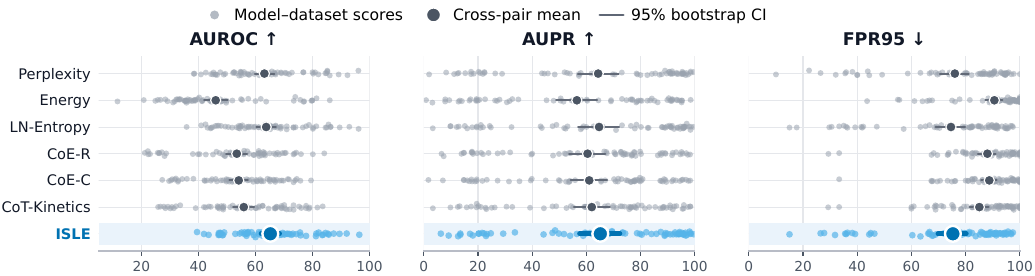}
\caption{
\textbf{Cross-task outcome separability of $S_{\mathrm{ISLE}}$.}
We report average AUROC, FPR95, and AUPR across MGSM, GPQA, HumanEval, and LLMs from different families and scales.
Higher AUROC/AUPR and lower FPR95 indicate stronger separation.
Full results are in Appendix~\ref{appendix:outcome_separability}.
}
    \label{fig:outcome_separability}
\end{figure*}

\subsection{Main Results}
\label{subsec:main_results}

We evaluate \textsc{GLIDE} on three benchmarks covering distinct forms of agent inference: HotpotQA~\citep{yang2018hotpotqa} for multi-hop reasoning, WebShop~\citep{yao2022webshop} for goal-directed sequential decision making, and Game of 24~\citep{tot} for symbolic arithmetic reasoning. HotpotQA is evaluated by \textbf{Exact Match (EM)}, WebShop by \textbf{Success Rate (SR)} and \textbf{Average Score}, and Game of 24 by \textbf{Success Rate}. We follow the task-specific evaluation protocols and report baseline details in Appendix~\ref{appendix:baseline_details}.

Tables~\ref{tab:hotpotqa}, \ref{tab:webshop_results}, and~\ref{tab:mbpp_results} show that \textsc{GLIDE} achieves strong performance across all three settings. Under the same open-source heterogeneous-agent configuration, \textsc{GLIDE} consistently improves over SYMPHONY-S, indicating that the gain comes from more reliable candidate evaluation and computation allocation rather than from a stronger model pool. The improvement is observed across qualitatively different sources of difficulty, including evidence selection in multi-hop reasoning, goal maintenance in long-horizon interaction, and branch isolation in symbolic arithmetic reasoning.

Compared with broader task-specific baselines, \textsc{GLIDE} remains competitive while using only lightweight intrinsic evaluation. This suggests that calibrated intrinsic rewards and adaptive branching jointly improve candidate comparison and computation allocation in heterogeneous agents.

\subsection{Preference Quality of Calibrated Rewards}
\label{subsec:preference_quality}
To verify that \textsc{GLIDE}'s task-level gains come from better candidate selection, we evaluate $R_{\mathrm{GLIDE}}$ on expanded child nodes from the three main tasks. Candidates are ranked by different signals and labeled by the final outcome of their continuations, testing whether successful continuations are prioritized. We report P@1, MRR, NDCG@3, per-candidate latency, and memory overhead.

Table~\ref{tab:discriminative_metrics_full} shows that $R_{\mathrm{GLIDE}}$ achieves the strongest overall ranking quality, outperforming probability-based intrinsic metrics, hidden-state baselines, extrinsic PRMs, and heterogeneous judge scores without task-specific supervision or additional verifier calls. Its gains over raw or uncalibrated signals show that distributional calibration turns $S_{\mathrm{ISLE}}$ from  intrinsic evidence into a more reliable heterogeneous candidate-comparison signal.

\begin{table}[t]
    \centering
     \caption{\textbf{Ablation Study} (Full Pool; WebShop: SR).}
    \label{tab:ablation_wrap}
    \begin{tabular}{lccc}
        \toprule
        \textbf{Variant} & \textbf{Hot} & \textbf{Web} & \textbf{G24} \\
        \midrule
        w/o Adap. Branch. & 0.68 & 0.50 & 0.52 \\
        Raw $S_{\mathrm{ISLE}}$ & 0.63 & 0.52 & 0.47 \\
        Calib. Judge & 0.66 & 0.56 & 0.50 \\
        Rand. Agent & 0.59 & 0.56 & 0.43 \\
        UCB Dispatch & 0.65 & 0.55 & 0.46 \\
        \rowcolor{gray!10} \textbf{Full} & \textbf{0.73} & \textbf{0.57} & \textbf{0.54} \\
        \bottomrule
    \end{tabular}
\end{table}

\subsection{Controlled Comparison and Ablation Study}
\label{subsec:controlled_ablation}

We further examine whether \textsc{GLIDE}'s gains hold under controlled agent-pool configurations and which components contribute to them. Table~\ref{tab:main_comparison_full} compares \textsc{GLIDE} with SYMPHONY using single-agent, dual-agent, and full heterogeneous pools. Across these settings, \textsc{GLIDE} improves task performance and token efficiency, showing that the gains do not come from stronger base models but from calibrated intrinsic evaluation and adaptive computation allocation. The results also indicate that adding more agents alone is insufficient, since heterogeneous candidates can introduce noisy branch comparison without calibrated score alignment.

Table~\ref{tab:ablation_wrap} isolates the effect of each component. Replacing $R_{\mathrm{GLIDE}}$ with raw $S_{\mathrm{ISLE}}$ weakens performance, confirming the need for distributional calibration. Removing adaptive branching reduces the benefit of uncertainty-aware computation allocation, while the calibrated judge variant shows that calibration alone is insufficient when the underlying signal comes from external judgment rather than generator-internal evidence. Alternative dispatching strategies remain competitive in some cases but do not match the full model. These results support the full design of \textsc{GLIDE}, where intrinsic evidence, agent-wise calibration, and adaptive branching work jointly within one controller. Appendix~\ref{appendix:main_supplementary_analyses} further reports test-time compute scaling, parameter sensitivity, calibration-window analysis, fixed-search evaluation baselines, and failure cases.

\subsection{Cross-Task Outcome Separability}
\label{subsec:outcome_separability}

Finally, we test whether $S_{\mathrm{ISLE}}$ remains outcome-relevant beyond the main agent-control setting. We evaluate multiple open-source LLMs on MGSM~\citep{data:mgsm}, GPQA~\citep{data:gpqa}, and HumanEval~\citep{data:humaneval}, covering multilingual mathematical reasoning, scientific question answering, and code generation. For each model--task pair, generated solutions are labeled by final task outcome, and each scoring method is evaluated by its ability to separate successful generations from failed ones. We report AUROC and AUPR for separability and FPR95 for false positives at high recall.

Figure~\ref{fig:outcome_separability} shows that $S_{\mathrm{ISLE}}$ maintains clear separability across models and tasks. This complements the step-level preference analysis by showing that the intrinsic signal is not only useful inside the calibrated control pipeline, but also remains associated with final task success under direct generation. Full results, layer-subset analysis, and memory-overhead evaluation are provided in Appendix~\ref{appendix:outcome_separability_details}.

\section{Conclusion}
\label{sec:conclusion}

This paper introduces \textsc{GLIDE}, which transforms layer-wise residual coherence into a cross-agent calibrated reward for heterogeneous LLM-agent search. 
By aligning intrinsic step evidence with agent-specific score distributions, \textsc{GLIDE} enables verifier-free value estimation and adaptive branching for efficient inference-time control.
Experiments show that this calibrated intrinsic reward improves task performance and step-level preference quality with low overhead, highlighting the promise of intrinsic signals for complex reasoning.

\section*{Acknowledgements}
This work is supported by National Natural Science Foundation of China (62666075), the Key Project of Fundamental Research of Yunnan Province (202401AS070138), the Program of Yunnan Key Laboratory of Intelligent Systems and Computing (202549CE340006), the Yunnan Fundamental Research Project (202501AT070231), and the Yunnan University Medical Research Foundation (K204209250001).

\section*{Limitations}
GLIDE requires access to hidden states, which may not be available for closed-source models. Its calibration is also designed for settings where a heterogeneous model pool repeatedly contributes candidate reasoning steps, so a single black-box model deployment is outside the primary scope of this work. Future work should study how similar calibration signals can be approximated when only partial model traces or API-level confidence signals are available.

\bibliography{glide}
\clearpage
\appendix

\begin{center}
    \textbf{\Large APPENDIX}
\end{center}
\hrule
\vspace{0.8em}

\begingroup
\small
\setlength{\tabcolsep}{0pt}
\renewcommand{\arraystretch}{1.06}

\begin{center}
\begin{tabularx}{0.94\linewidth}{@{}>{\raggedright\arraybackslash}X@{\hspace{1em}}r@{}}

    \textbf{A \quad GLIDE Algorithm} 
    & \pageref{appendix:algorithm} \\[0.25em]

    \textbf{B \quad Theoretical Background} 
    & \pageref{sec:appendix_background} \\
    \quad B.1 Markov Decision Processes and LLM Agents 
    & \pageref{subsec:mdp_llm} \\
    \quad B.2 Monte Carlo Tree Search 
    & \pageref{subsec:mcts_algorithm} \\[0.25em]

    \textbf{C \quad Theoretical Analysis} 
    & \pageref{sec:theoretical_analysis} \\
    \quad C.1 Setup and Notation
    & \pageref{app:theory_setup} \\
    \quad C.2 Analysis of Assumption~\ref{assump:coherent_residual_accumulation}
    & \pageref{app:coherent_residual_accumulation} \\
    \quad C.2.1 Low-dimensional Residual Subspace 
    & \pageref{app:residual_subspace} \\
    \quad C.2.2 Positive Local-to-global Coherence Margin 
    & \pageref{app:coherence_margin} \\
    \quad C.3 Proof of Proposition~\ref{prop:preference_consistency}
    & \pageref{app:preference_consistency} \\[0.25em]

    \textbf{D \quad Main Experimental Details} 
    & \pageref{appendix:main_experiment_details} \\
    \quad D.1 Infrastructure and Model Configuration
    & \pageref{app:main_model_config} \\
    \quad D.2 MCTS and GLIDE Hyperparameters
    & \pageref{app:main_hyperparams} \\
    \quad D.3 Task-Specific Prompting and Settings
    & \pageref{app:task_settings} \\
    \quad D.4 Main Benchmark Baselines
    & \pageref{appendix:baseline_details} \\[0.25em]

    \textbf{E \quad Supplementary Analyses for Main Experiments} 
    & \pageref{appendix:main_supplementary_analyses} \\
    \quad E.1 Test-Time Compute Scaling 
    & \pageref{appendix:cost} \\
    \quad E.2 Algorithmic Parameter Analysis
    & \pageref{appendix:parameter_analysis} \\
    \quad E.3 MCTS Evaluation Baselines under Fixed Search
    & \pageref{app:mcts_eval_baselines} \\
    \quad E.4 Step-Level Preference and Efficiency Analysis
    & \pageref{app:step_preference_details} \\
    \quad E.5 Failure Case Analysis 
    & \pageref{app:failure_case_analysis} \\[0.25em]

    \textbf{F \quad Cross-Task Outcome Separability} 
    & \pageref{appendix:outcome_separability_details} \\
    \quad F.1 Datasets and Models
    & \pageref{app:outcome_model_config} \\
    \quad F.2 Generation Protocol 
    & \pageref{app:outcome_generation_protocol} \\
    \quad F.3 Outcome-Separability Results 
    & \pageref{appendix:outcome_separability} \\
    \quad F.4 Layer-Subset Analysis
    & \pageref{app:layer_subset_analysis} \\
    \quad F.5 Memory Overhead of Hidden-State Scoring
    & \pageref{app:vram_overhead} \\[0.25em]

    \textbf{G \quad Use of AI Assistants} 
    & \pageref{useai} \\

    \textbf{H \quad Artifacts Statements} 
    & \pageref{app:artifacts} \\

\end{tabularx}
\end{center}

\endgroup

\vspace{0.8em}
\hrule
\vspace{1em}

\section{Algorithm Pseudocode}
\label{appendix:algorithm}

Algorithm~\ref{alg:glide} presents the pseudocode of the proposed GLIDE framework.

\begin{algorithm}[t]
   \small
   \caption{\textsc{GLIDE}: Intrinsic Reward-Guided Heterogeneous MCTS}
   \label{alg:glide}
\begin{algorithmic}[1]
   \STATE {\bfseries Input:} initial state $s_0$, agent pool $\mathcal{M}=\{M_i\}_{i=1}^{B}$, MCTS budget $K$, UCT constant $c$, branching bounds $n_{\min},n_{\max}$.
   \STATE {\bfseries Initialize:} tree $\mathcal{T}\leftarrow\{s_0\}$, node capacity $b(s_0)\leftarrow n_{\max}$, and FIFO windows $\mathcal{W}_i^{S},\mathcal{W}_i^{H}$ for each $M_i\in\mathcal{M}$.

   \FOR{$k=1$ {\bfseries to} $K$}
      \STATE $s\leftarrow s_0$

      \STATE \COMMENT{\textbf{1. Selection}}
      \WHILE{$s$ is not terminal \AND $|\mathrm{ch}(s)|\ge b(s)$}
         \STATE $s\leftarrow \operatorname*{argmax}_{s'\in\mathrm{ch}(s)} \mathrm{UCT}(s')$ \COMMENT{Eq.~\eqref{eq:uct}}
      \ENDWHILE

      \STATE \COMMENT{\textbf{2. Expansion and intrinsic evaluation}}
      \WHILE{$s$ is not terminal \AND $|\mathrm{ch}(s)|< b(s)$}
         \STATE Sample agent $M_i\sim\mathrm{Uniform}(\mathcal{M})$.
         \STATE Generate step $a$ from state $s$ using $M_i$, yielding successor $s'$.
         \STATE Extract hidden states and token distributions from the generation pass.

         \STATE Compute $S_{\mathrm{ISLE}}^{(i)}(s,a)$ using Eq.~\eqref{eq:isle_score}.
         \STATE Compute $S_{\mathrm{rank}}^{(i)}(s,a)$ from $\mathcal{W}_i^{S}$ using Eq.~\eqref{eq:rank_score}.
         \STATE Compute $R_{\mathrm{GLIDE}}^{(i)}(s,a)$ using Eq.~\eqref{eq:glide_reward}.

         \STATE Compute mean entropy $\bar{\mathcal{H}}_i(s,a)$ and normalized uncertainty $\widetilde{\mathcal{H}}_i(s,a)$ using Eq.~\eqref{eq:relative_entropy}.
         \STATE Set successor capacity $b(s')\leftarrow n(s,a)$ using Eq.~\eqref{eq:adaptive_branching}.

         \STATE Add $s'$ to $\mathcal{T}$ as a child of $s$ and store $R_{\mathrm{GLIDE}}^{(i)}(s,a)$.
         \STATE Backpropagate using Eq.~\eqref{eq:back}.

         \STATE Insert $S_{\mathrm{ISLE}}^{(i)}(s,a)$ into $\mathcal{W}_i^{S}$ and $\bar{\mathcal{H}}_i(s,a)$ into $\mathcal{W}_i^{H}$.
      \ENDWHILE
   \ENDFOR

   \STATE {\bfseries Return} the best reasoning path from $\mathcal{T}$.
\end{algorithmic}
\end{algorithm}

\section{Theoretical Background}
\label{sec:appendix_background}

\subsection{Markov Decision Processes and LLM Agents}
\label{subsec:mdp_llm}

The sequential decision-making problem is formally defined by the Markov Decision Process (MDP) tuple $(\mathcal{S}, \mathcal{A}, \mathcal{P}, \mathcal{R}, \gamma)$~\cite{bellman1966MDP}. Here, $\mathcal{S}$ denotes the state space, $\mathcal{A}$ represents the action space, $\mathcal{P}: \mathcal{S} \times \mathcal{A} \rightarrow \Delta(\mathcal{S})$ defines the transition dynamics, $\mathcal{R}: \mathcal{S} \times \mathcal{A} \rightarrow \mathbb{R}$ is the reward function, and $\gamma \in [0,1]$ serves as the discount factor. At each timestep $t$, the agent observes a state $s_t \in \mathcal{S}$, selects an action $a_t \in \mathcal{A}$, transitions to a new state $s_{t+1} \sim \mathcal{P}(\cdot \mid s_t, a_t)$, and receives a scalar reward $r_t = \mathcal{R}(s_t, a_t)$. The objective is to find a policy $\pi$ that maximizes the expected cumulative discounted return $\mathbb{E}[\sum_{t=0}^{\infty} \gamma^t r_t]$.

Large Language Models (LLMs) can be naturally integrated into this framework to support high-level reasoning. Specifically, an LLM can serve as a \textit{policy} by generating actions (tokens or reasoning steps) conditioned on language-based state representations; as a \textit{value function} by estimating expected returns from textual trajectories; or as a \textit{world model} by predicting future states and rewards through learned internal knowledge. Unlike traditional reinforcement learning agents that rely on explicit environment modeling and manually designed signals, LLM-based agents leverage pretraining on large corpora to internalize commonsense, domain knowledge, and structured reasoning. This allows them to operate effectively in complex, open-ended environments with minimal task-specific engineering.

\subsection{Monte Carlo Tree Search (MCTS)}
\label{subsec:mcts_algorithm}

Monte Carlo Tree Search (MCTS)~\cite{coulom2006MCTS} is a sample-based planning algorithm that balances exploration (trying under-sampled actions) and exploitation (refining high-reward actions) through iterative tree search. When integrated with LLM-based agents, MCTS leverages the language model's prior knowledge to guide efficient exploration in sequential decision making.

Formally, given the MDP setup described above, MCTS incrementally builds a partial search tree rooted at the initial state $s_0$. Each node in the tree represents a state $s$, and edges represent actions $a$. The algorithm maintains empirical statistics for each node-action pair: the visit count $N(s, a)$ and the estimated mean action value $Q(s, a)$. The search process repeats the following four phases until a computational budget is exhausted:

\paragraph{1. Selection.}
Starting from the root, the algorithm traverses the tree by recursively selecting the optimal child node. This step typically employs the Upper Confidence Bound for Trees (UCT) criterion to handle the exploration-exploitation trade-off:
\begin{equation}
\begin{aligned}
    a_t
    &=
    \operatorname*{argmax}_{a \in \mathcal{A}(s_t)}
    \left[
    Q(s_t, a)
    + c \sqrt{\frac{\ln N(s_t)}{N(s_t, a)}}
    \right],
\end{aligned}
    \label{eq:uct}
\end{equation}
where $N(s_t) = \sum_{a} N(s_t, a)$ is the total visit count of the parent node, and $c > 0$ is the exploration constant that controls the degree of exploration.

\paragraph{2. Expansion.}
Upon reaching a leaf node $s_L$ that is not terminal and has unvisited children, the tree is expanded by instantiating a new child node. In the context of LLMs, this corresponds to sampling a new token or reasoning step $a$ from the policy $\pi$ and appending the resulting state $s' \sim \mathcal{P}(\cdot \mid s_L, a)$ to the tree.

\paragraph{3. Simulation (Rollout).}
From the newly expanded node $s'$, a simulation is executed to estimate its potential value. This typically involves generating a full trajectory of length $T$ using a lightweight rollout policy $\pi_{\text{rollout}}$ until a terminal state is reached:
\begin{equation}
\begin{aligned}
    R_{\text{sim}}
    &=
    \sum_{t=0}^{T} \gamma^{t}\mathcal{R}(s_t,a_t),\\
    a_t
    &\sim
    \pi_{\mathrm{rollout}}(\cdot\mid s_t).
\end{aligned}
\end{equation}
The cumulative reward $R_{\text{sim}}$ serves as a Monte Carlo estimate of the state value $V(s')$. In recent LLM reasoning frameworks, this costly rollout is sometimes substituted by a learned value function or heuristic evaluation.

\paragraph{4. Backpropagation.}
The estimated value $R_{\text{sim}}$ is propagated backward from the leaf $s'$ up to the root. For every edge $(s, a)$ along the traversal path, the statistics are updated via incremental averaging:
\begin{equation}
\begin{aligned}
\label{eq:back}
    N(s, a) &\leftarrow N(s, a) + 1, \\
    Q(s, a) &\leftarrow Q(s, a) + \frac{R_{\text{sim}} - Q(s, a)}{N(s, a)}.
\end{aligned}
\end{equation}
Under the assumption that every action is eventually explored infinitely often, the UCT update guarantees that $Q(s, a)$ converges to the optimal value function, guiding the agent toward the optimal policy.


\section{Theoretical Analysis}
\label{sec:theoretical_analysis}

This appendix provides detailed derivations for the theoretical claims used in Section~\ref{subsec:irc_score_theory} and Section~\ref{subsec:calibration}.
The layer-wise analysis is local to a fixed transition $(x,a)$, and the dependence on $(x,a)$ is omitted unless otherwise stated.
The preference-consistency analysis fixes an agent and a search state, and compares candidate steps generated from that state.

\subsection{Setup and Notation}
\label{app:theory_setup}
Let $\mathbf{h}^{(l)}\in\mathbb{R}^{d}$ denote the pooled hidden representation of a candidate step at layer $l$, where $l=0,\dots,L$.
The local residual update is
\begin{equation}
\mathbf{u}_l
=
\mathbf{h}^{(l+1)}
-
\mathbf{h}^{(l)},
\qquad
l=0,\dots,L-1 .
\label{eq:app_local_update}
\end{equation}
The global residual update is
\begin{equation}
\mathbf{u}_{\mathrm{global}}
=
\mathbf{h}^{(L)}
-
\mathbf{h}^{(0)}
=
\sum_{l=0}^{L-1}
\mathbf{u}_l .
\label{eq:app_global_update}
\end{equation}
The residual-coherence term $\Phi_{\mathrm{coh}}$ in Eq.~\eqref{eq:coherence_score} is based on the accumulated local-to-global agreement
$\frac{1}{L}\sum_{l=0}^{L-1}\cos(\mathbf{u}_l,\mathbf{u}_{\mathrm{global}})$.

\subsection{Analysis of Assumption~\ref{assump:coherent_residual_accumulation}}
\label{app:coherent_residual_accumulation}

Assumption~\ref{assump:coherent_residual_accumulation} states that, for a fixed candidate step, layer-wise residual updates concentrate near a shared low-dimensional residual subspace, within which a dominant component accumulates consistently across layers.
We analyze these two implications separately.
The shared subspace yields an approximate low-rank structure, while the stable dominant component yields a positive local-to-global coherence margin.

\subsubsection{Low-dimensional residual subspace}
\label{app:residual_subspace}

A sufficient formalization of the shared-subspace part of Assumption~\ref{assump:coherent_residual_accumulation} is
\begin{equation}
\mathbf{u}_l
=
\mathbf{G}\boldsymbol{\zeta}_l
+
\boldsymbol{\epsilon}_l,
\qquad
l=0,\dots,L-1 ,
\label{eq:app_factorized_update}
\end{equation}
where $\mathbf{G}\in\mathbb{R}^{d\times k}$ spans a step-local residual subspace, $\boldsymbol{\zeta}_l\in\mathbb{R}^{k}$ are layer-specific coefficients, $k<\min\{d,L\}$, and $\|\boldsymbol{\epsilon}_l\|_2\le\delta$.

Let
\begin{equation}
\mathbf{U}
=
[
\mathbf{u}_0,\dots,\mathbf{u}_{L-1}
]
\in
\mathbb{R}^{d\times L}
\label{eq:app_update_matrix}
\end{equation}
collect the layer-wise residual updates, and define
\begin{align}
\mathbf{Z}
&=
[
\boldsymbol{\zeta}_0,\dots,
\boldsymbol{\zeta}_{L-1}
]
\in
\mathbb{R}^{k\times L},
\label{eq:app_factor_coeff_matrix}
\\
\mathbf{E}
&=
[
\boldsymbol{\epsilon}_0,\dots,
\boldsymbol{\epsilon}_{L-1}
]
\in
\mathbb{R}^{d\times L}.
\label{eq:app_error_matrix}
\end{align}
Stacking Eq.~\eqref{eq:app_factorized_update} over layers gives
\begin{equation}
\mathbf{U}
=
\mathbf{G}\mathbf{Z}
+
\mathbf{E}.
\label{eq:app_U_decomposition}
\end{equation}

If $\boldsymbol{\epsilon}_l=\mathbf{0}$ for all $l$, then $\mathbf{U}=\mathbf{G}\mathbf{Z}$ and
\begin{equation}
\begin{aligned}
\operatorname{rank}(\mathbf{U})
&=
\operatorname{rank}(\mathbf{G}\mathbf{Z})
\\
&\le
\min\{
\operatorname{rank}(\mathbf{G}),
\operatorname{rank}(\mathbf{Z})
\}
\\
&\le
k .
\end{aligned}
\label{eq:app_rank_chain}
\end{equation}
Thus, in the noiseless case, the layer-wise residual updates lie in a subspace of rank at most $k$.

With bounded perturbations, $\operatorname{rank}(\mathbf{G}\mathbf{Z})\le k$.
By the variational characterization of singular values,
\begin{align}
\sigma_{k+1}(\mathbf{U})
&=
\min_{\operatorname{rank}(\mathbf{A})\le k}
\|
\mathbf{U}-\mathbf{A}
\|_2
\nonumber\\
&\le
\|
\mathbf{U}-\mathbf{G}\mathbf{Z}
\|_2
=
\|\mathbf{E}\|_2 .
\label{eq:app_singular_variational}
\end{align}
Since the spectral norm is bounded by the Frobenius norm,
\begin{align}
\|\mathbf{E}\|_2
&\le
\|\mathbf{E}\|_F
\nonumber\\
&=
\left(
\sum_{l=0}^{L-1}
\|\boldsymbol{\epsilon}_l\|_2^2
\right)^{1/2}
\nonumber\\
&\le
\sqrt{L}\,\delta .
\label{eq:app_E_norm_bound}
\end{align}
Combining Eq.~\eqref{eq:app_singular_variational} and Eq.~\eqref{eq:app_E_norm_bound} gives
\begin{equation}
\sigma_{k+1}(\mathbf{U})
\le
\sqrt{L}\,\delta .
\label{eq:app_low_rank_bound}
\end{equation}

This shows that the residual-update matrix remains close to a low-rank structure whenever the perturbations around the shared residual subspace are bounded.
It formalizes the first part of Assumption~\ref{assump:coherent_residual_accumulation}: layer-wise residual updates are not arbitrary high-dimensional fluctuations, but concentrate near a common low-dimensional structure.

\subsubsection{Positive local-to-global coherence margin}
\label{app:coherence_margin}

We next formalize the stable-accumulation part of Assumption~\ref{assump:coherent_residual_accumulation}.
A sufficient condition is that the residual updates contain a shared dominant direction:
\begin{equation}
\mathbf{u}_l
=
\eta_l\mathbf{g}
+
\boldsymbol{\epsilon}_l,
\qquad
l=0,\dots,L-1 ,
\label{eq:app_dominant_factor}
\end{equation}
where $\|\mathbf{g}\|_2=1$, $0<\eta_{\min}\le\eta_l\le\eta_{\max}$, and $\|\boldsymbol{\epsilon}_l\|_2\le\delta$.
The dominant direction $\mathbf{g}$ can be viewed as a stable component inside the shared residual subspace.

Define the average global update
\begin{equation}
\bar{\mathbf{u}}
=
\frac{1}{L}
\mathbf{u}_{\mathrm{global}}
=
\bar{\eta}\mathbf{g}
+
\bar{\boldsymbol{\epsilon}},
\label{eq:app_average_update}
\end{equation}
where
\begin{equation}
\bar{\eta}
=
\frac{1}{L}
\sum_{l=0}^{L-1}
\eta_l,
\qquad
\bar{\boldsymbol{\epsilon}}
=
\frac{1}{L}
\sum_{l=0}^{L-1}
\boldsymbol{\epsilon}_l .
\label{eq:app_average_terms}
\end{equation}
By construction,
\begin{equation}
\eta_{\min}
\le
\bar{\eta}
\le
\eta_{\max},
\qquad
\|\bar{\boldsymbol{\epsilon}}\|_2
\le
\delta .
\label{eq:app_average_bounds}
\end{equation}

Assume that the stable component dominates the perturbation scale:
\begin{equation}
\eta_{\min}^{2}
>
2\eta_{\max}\delta+\delta^2 .
\label{eq:app_positive_margin_condition}
\end{equation}
This condition implies $\eta_{\min}>\delta$, so $\mathbf{u}_l$ and $\bar{\mathbf{u}}$ are nonzero and the cosine similarities are well-defined.
We show that
\begin{equation}
\begin{aligned}
\frac{1}{L}
\sum_{l=0}^{L-1}
\cos(
\mathbf{u}_l,
\mathbf{u}_{\mathrm{global}}
)
\ge
\frac{
\eta_{\min}^{2}
-
2\eta_{\max}\delta
-
\delta^2
}{
(\eta_{\max}+\delta)^2
}
>0 .
\end{aligned}
\label{eq:app_coherence_margin}
\end{equation}

Because cosine similarity is invariant to positive rescaling of the second argument,
\begin{equation}
\cos(
\mathbf{u}_l,
\mathbf{u}_{\mathrm{global}}
)
=
\cos(
\mathbf{u}_l,
\bar{\mathbf{u}}
),
\label{eq:app_cos_rescale}
\end{equation}
where $\bar{\mathbf{u}}=\mathbf{u}_{\mathrm{global}}/L$.
For each layer $l$, expand the numerator:
\begin{align}
\left\langle
\mathbf{u}_l,
\bar{\mathbf{u}}
\right\rangle
&=
\left\langle
\eta_l\mathbf{g}
+
\boldsymbol{\epsilon}_l,
\bar{\eta}\mathbf{g}
+
\bar{\boldsymbol{\epsilon}}
\right\rangle
\nonumber\\
&=
\eta_l\bar{\eta}
+
\eta_l
\langle
\mathbf{g},
\bar{\boldsymbol{\epsilon}}
\rangle
\nonumber\\
&\quad
+
\bar{\eta}
\langle
\boldsymbol{\epsilon}_l,
\mathbf{g}
\rangle
+
\langle
\boldsymbol{\epsilon}_l,
\bar{\boldsymbol{\epsilon}}
\rangle .
\label{eq:app_inner_expansion}
\end{align}
Using Cauchy--Schwarz and Eq.~\eqref{eq:app_average_bounds},
\begin{align}
\left\langle
\mathbf{u}_l,
\bar{\mathbf{u}}
\right\rangle
&\ge
\eta_l\bar{\eta}
-
\eta_l
\|\bar{\boldsymbol{\epsilon}}\|_2
-
\bar{\eta}
\|\boldsymbol{\epsilon}_l\|_2
\nonumber\\
&\quad
-
\|\boldsymbol{\epsilon}_l\|_2
\|\bar{\boldsymbol{\epsilon}}\|_2
\nonumber\\
&\ge
\eta_{\min}^{2}
-
2\eta_{\max}\delta
-
\delta^2 .
\label{eq:app_numerator_bound}
\end{align}
Under Eq.~\eqref{eq:app_positive_margin_condition}, this lower bound is strictly positive.

The denominator is upper bounded by
\begin{align}
\|\mathbf{u}_l\|_2
&=
\|
\eta_l\mathbf{g}
+
\boldsymbol{\epsilon}_l
\|_2
\nonumber\\
&\le
\eta_l+\|\boldsymbol{\epsilon}_l\|_2
\le
\eta_{\max}+\delta ,
\label{eq:app_local_norm_bound}
\\
\|\bar{\mathbf{u}}\|_2
&=
\|
\bar{\eta}\mathbf{g}
+
\bar{\boldsymbol{\epsilon}}
\|_2
\nonumber\\
&\le
\bar{\eta}
+
\|\bar{\boldsymbol{\epsilon}}\|_2
\le
\eta_{\max}+\delta .
\label{eq:app_average_norm_bound}
\end{align}
Thus,
\begin{equation}
\|\mathbf{u}_l\|_2
\,
\|\bar{\mathbf{u}}\|_2
\le
(\eta_{\max}+\delta)^2 .
\label{eq:app_denominator_bound}
\end{equation}
Combining Eq.~\eqref{eq:app_numerator_bound} and Eq.~\eqref{eq:app_denominator_bound}, for every layer $l$,
\begin{align}
\cos(
\mathbf{u}_l,
\mathbf{u}_{\mathrm{global}}
)
&=
\cos(
\mathbf{u}_l,
\bar{\mathbf{u}}
)
\nonumber\\
&\ge
\frac{
\eta_{\min}^{2}
-
2\eta_{\max}\delta
-
\delta^2
}{
(\eta_{\max}+\delta)^2
}.
\label{eq:app_layer_cos_bound}
\end{align}
Averaging Eq.~\eqref{eq:app_layer_cos_bound} over layers gives Eq.~\eqref{eq:app_coherence_margin}.

This analysis formalizes the second part of Assumption~\ref{assump:coherent_residual_accumulation}: when a stable component dominates the perturbation scale, local residual updates have positive average agreement with the global residual update.
This supports the residual-coherence term $\Phi_{\mathrm{coh}}$ in Eq.~\eqref{eq:coherence_score} and, together with the decisiveness gate $\Phi_{\mathrm{dec}}$, the intrinsic stability score $S_{\mathrm{ISLE}}$ in Eq.~\eqref{eq:isle_score}.

\subsection{Proof of Proposition~\ref{prop:preference_consistency}}
\label{app:preference_consistency}

We provide the detailed derivation of Proposition~\ref{prop:preference_consistency}.
Fix an agent $M_i$ and a search state $x$.
Consider two candidate steps $a$ and $a'$ generated by $M_i$ at state $x$.
Let
\begin{equation}
\Delta_q(a,a')
=
q(x,a)-q(x,a')
\label{eq:app_delta_q_def}
\end{equation}
denote the latent utility gap, and let
\begin{equation}
\Delta_S^{(i)}(a,a')
=
S_{\mathrm{ISLE}}^{(i)}(x,a)
-
S_{\mathrm{ISLE}}^{(i)}(x,a')
\label{eq:app_delta_s_def}
\end{equation}
denote the intrinsic-score gap.

We use the following sufficient formulation of the noisy monotone surrogate condition.
For any candidate step $b$ generated by agent $M_i$ at state $x$,
\begin{equation}
S_{\mathrm{ISLE}}^{(i)}(x,b)
=
\mu_i(q(x,b))
+
\xi_b ,
\label{eq:app_noisy_surrogate}
\end{equation}
where $\mu_i$ is order-preserving with margin $\lambda_i>0$:
\begin{equation}
\begin{aligned}
q(x,a)>q(x,a')
\quad\Rightarrow\quad
&
\mu_i(q(x,a))
-
\mu_i(q(x,a'))
\\
&\ge
\lambda_i
\Delta_q(a,a') .
\end{aligned}
\label{eq:app_monotone_gap_condition}
\end{equation}
We further assume that the noise difference
\[
Z=\xi_{a'}-\xi_a
\]
is mean-zero and $\tau_i$-sub-Gaussian:
\begin{equation}
\mathbb{E}
\exp(\nu Z)
\le
\exp\left(
\frac{\tau_i^{2}\nu^{2}}{2}
\right),
\qquad
\forall \nu\in\mathbb{R}.
\label{eq:app_difference_subgaussian_direct}
\end{equation}
This condition holds, for example, when $\xi_a$ and $\xi_{a'}$ are independent, mean-zero, and $\sigma_i$-sub-Gaussian, in which case $\tau_i=\sqrt{2}\sigma_i$.

Assume $\Delta_q(a,a')>0$.
The event $\Delta_S^{(i)}(a,a')\le 0$ implies
\begin{align}
S_{\mathrm{ISLE}}^{(i)}(x,a)
&\le
S_{\mathrm{ISLE}}^{(i)}(x,a'),
\nonumber\\
\xi_{a'}-\xi_a
&\ge
\mu_i(q(x,a))
-
\mu_i(q(x,a'))
\nonumber\\
&\ge
\lambda_i\Delta_q(a,a') .
\label{eq:app_misrank_implies_noise}
\end{align}
Therefore,
\begin{equation}
\Pr\!\left(
\Delta_S^{(i)}(a,a')\le 0
\right)
\le
\Pr\!\left(
Z
\ge
\lambda_i\Delta_q(a,a')
\right).
\label{eq:app_misrank_to_tail}
\end{equation}
Applying the one-sided sub-Gaussian tail bound gives
\begin{equation}
\Pr\!\left(
Z
\ge
\lambda_i\Delta_q(a,a')
\right)
\le
\exp\left(
-\frac{
\lambda_i^2
\Delta_q(a,a')^{2}
}{
2\tau_i^{2}
}
\right).
\label{eq:app_tail_bound}
\end{equation}
Combining Eq.~\eqref{eq:app_misrank_to_tail} and Eq.~\eqref{eq:app_tail_bound} yields
\begin{equation}
\Pr\!\left(
\Delta_S^{(i)}(a,a')\le 0
\right)
\le
\exp\left(
-\frac{
\lambda_i^2
\Delta_q(a,a')^{2}
}{
2\tau_i^{2}
}
\right).
\label{eq:app_ranking_bound}
\end{equation}
This matches Eq.~\eqref{eq:preference_consistency_bound} with
\begin{equation}
\kappa_i
=
\frac{\lambda_i^2}{2\tau_i^2}.
\label{eq:app_kappa_def}
\end{equation}
Under the independent $\sigma_i$-sub-Gaussian sufficient condition above, this reduces to
\[
\kappa_i
=
\frac{\lambda_i^2}{4\sigma_i^2}.
\]

\subsection{Summary}

The analysis is aligned with the main text.
Assumption~\ref{assump:coherent_residual_accumulation} implies that layer-wise residual updates remain close to a shared low-dimensional residual subspace and can accumulate through a stable dominant component.
The first implication yields the approximate low-rank bound in Eq.~\eqref{eq:app_low_rank_bound}, while the second yields the positive local-to-global coherence margin in Eq.~\eqref{eq:app_coherence_margin}.
These consequences justify the residual-coherence term $\Phi_{\mathrm{coh}}$ used by $S_{\mathrm{ISLE}}$.
Proposition~\ref{prop:preference_consistency} further shows that, within each agent, the probability that $S_{\mathrm{ISLE}}$ reverses the order of two candidate steps decays exponentially with their latent utility gap.

These results do not claim that all Transformer computations exactly satisfy Assumption~\ref{assump:coherent_residual_accumulation}.
Rather, they identify a structural regime in which layer-wise residual coherence is mathematically meaningful as an intrinsic step-level preference signal for heterogeneous search.


\begin{table*}[t]
\centering
\caption{
\textbf{Test-time compute scaling on HotpotQA.}
We vary the MCTS budget \(K\) and report per-question averages.
Results for ToT, RAP, and LATS are taken from LATS~\citep{zhou2024language}, while SYMPHONY results are taken from SYMPHONY~\citep{symphony}.
\textsc{GLIDE} maintains strong performance while using fewer nodes and substantially fewer tokens than verifier-based search baselines at comparable or larger search budgets.
}
\label{tab:token_cost}
\small
\setlength{\tabcolsep}{6pt}
\begin{adjustbox}{max width=\textwidth}
\begin{tabular}{lccc}
\toprule
\textbf{Budget / Method} & \textbf{Exact Match} $\uparrow$ & \textbf{\# Nodes} $\downarrow$ & \textbf{Token Consumption} $\downarrow$ \\
\midrule
\multicolumn{4}{l}{\textbf{\(K=10\)}} \\
ToT~\citep{tot}             & 0.34 & 33.97 & -- \\
RAP~\citep{hao2023reasoning} & 0.44 & 31.53 & -- \\
LATS~\citep{zhou2024language} & 0.44 & 28.42 & -- \\
SYMPHONY-S~\citep{symphony}  & 0.59 & 16.39 & -- \\
SYMPHONY-L~\citep{symphony}  & \textbf{0.79} & \textbf{9.47} & 7,906 \\
\textsc{GLIDE}              & 0.73 & 15.74 & \textbf{6,052} \\
\midrule
\multicolumn{4}{l}{\textbf{\(K=30\)}} \\
ToT~\citep{tot}             & 0.39 & 47.54 & -- \\
RAP~\citep{hao2023reasoning} & 0.50 & 37.71 & -- \\
LATS~\citep{zhou2024language} & 0.52 & 34.12 & -- \\
\textsc{GLIDE}              & \textbf{0.76} & \textbf{26.21} & \textbf{9,872} \\
\midrule
\multicolumn{4}{l}{\textbf{\(K=50\)}} \\
ToT~\citep{tot}             & 0.49 & 84.05 & 210,215 \\
RAP~\citep{hao2023reasoning} & 0.54 & 70.60 & 176,500 \\
LATS~\citep{zhou2024language} & 0.61 & 66.65 & 173,290 \\
\textsc{GLIDE}              & \textbf{0.78} & \textbf{54.10} & \textbf{31,504} \\
\bottomrule
\end{tabular}
\end{adjustbox}
\end{table*}

\begin{figure*}[t]
    \centering
    \includegraphics[width=0.95\linewidth]{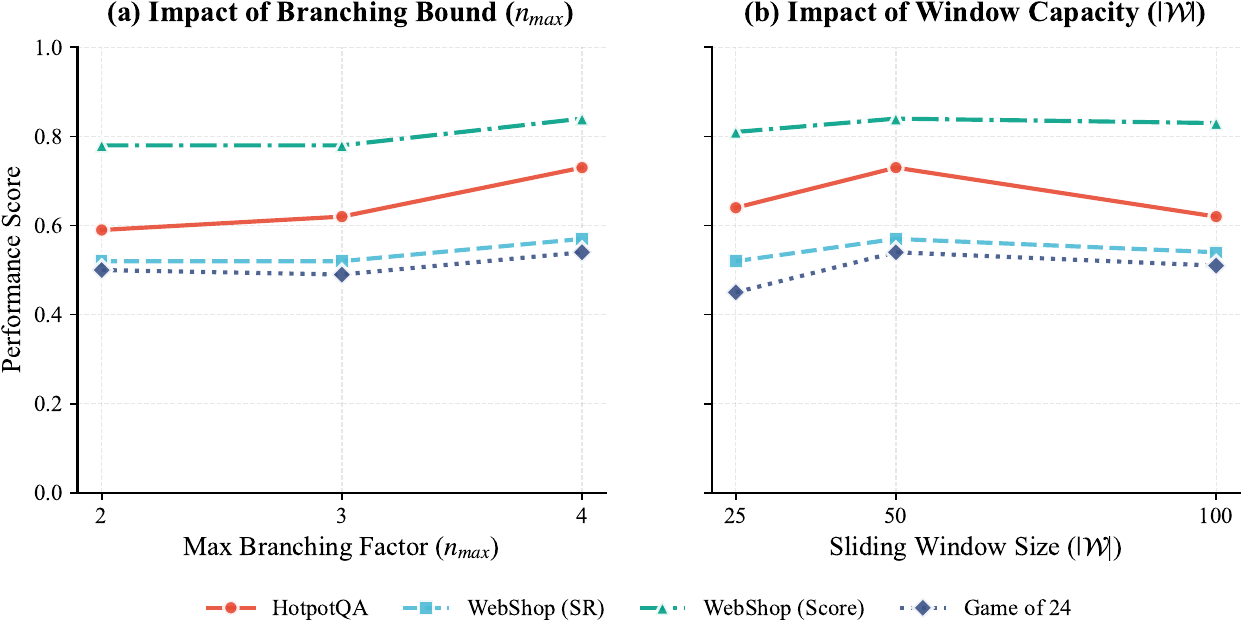}
\caption{\textbf{Hyperparameter Sensitivity Analysis.} 
Performance trends across three datasets (four metrics) under varying configurations.
\textbf{(a)} Impact of the search branching bound $n_{\max}$. The results show that performance remains generally stable across different branching bounds, with moderate gains obtained when allowing a slightly larger search space.
\textbf{(b)} Impact of the sliding window capacity $|\mathcal{W}|$. The sliding window balances statistical stability and adaptation to local distribution shifts. The results show a broad performance plateau around $|\mathcal{W}|=50$, indicating that this default setting robustly covers typical context shifts without task-specific tuning.}
\label{fig:ablation_study}
\end{figure*}


\section{Main Experimental Details}
\label{appendix:main_experiment_details}

This section provides implementation details for the main experiments on HotpotQA, WebShop, and Game of 24.
These experiments evaluate \textsc{GLIDE} within the MCTS-based heterogeneous agent setting and use the same open-source model pool as SYMPHONY-S for controlled comparison.

\subsection{Infrastructure and Model Configuration}
\label{app:main_model_config}

\paragraph{Model pool.}
Following SYMPHONY-S~\citep{symphony}, we use open-source models compatible with consumer-grade hardware:
Qwen2.5-7B-Instruct-1M~\citep{model:qwen},
Mistral-7B-Instruct-v0.3~\citep{jiang2023mistral7b},
and Llama-3.1-8B-Instruct~\citep{grattafiori2024llama3herdmodels}.
The same model pool is used for both \textsc{GLIDE} and the reproduced SYMPHONY-S baseline.

\paragraph{Compute environment.}
All main experiments are conducted on a local server with 3 NVIDIA RTX 4090 GPUs, each with 24GB VRAM.
This setup is used to evaluate whether heterogeneous search can achieve competitive performance under accessible compute constraints.

\subsection{MCTS and \textsc{GLIDE} Hyperparameters}
\label{app:main_hyperparams}

Unless otherwise stated, all main experiments follow a unified search and generation protocol.

\paragraph{MCTS parameters.}
The maximum number of MCTS iterations is set to $K=10$ across HotpotQA, WebShop, and Game of 24.
The UCT exploration constant is set to $c=\sqrt{2}$, following common MCTS practice and prior LLM-agent search settings~\citep{zhou2024language,symphony}.

\paragraph{\textsc{GLIDE} parameters.}
For distributional calibration, the score window and entropy window are both set to 50:
$|\mathcal{W}_i^{S}|=|\mathcal{W}_i^{H}|=50$.
For adaptive branching, we set $n_{\min}=1$ and $n_{\max}=4$ in Eq.~\eqref{eq:adaptive_branching}.
The upper bound $n_{\max}=4$ matches the fixed expansion width used by SYMPHONY, keeping the maximum branching budget comparable.

\paragraph{Generation parameters.}
We use temperature 0.2, nucleus sampling with top-$p=0.9$, and a maximum length of 200 tokens for each atomic reasoning step.

\subsection{Task-Specific Prompting and Settings}
\label{app:task_settings}

\paragraph{HotpotQA.}
HotpotQA is a multi-hop question answering benchmark requiring evidence aggregation and logical synthesis.
We use a 3-shot Chain-of-Thought prompting format, where demonstrations guide the model to decompose complex questions into intermediate reasoning steps.

\paragraph{WebShop.}
WebShop is a simulated e-commerce environment requiring goal-directed browser interactions.
We use a 1-shot interaction prompt to demonstrate the valid action space, including search, click, and purchase actions.

\paragraph{Game of 24.}
Game of 24 is a symbolic arithmetic task where the model must combine four numbers to reach 24.
We use a 2-shot prompting format and instruct the model to propose one atomic arithmetic operation at a time, providing explicit state transitions for MCTS expansion.

\subsection{Main Benchmark Baselines}
\label{appendix:baseline_details}

To rigorously evaluate the effectiveness of \textsc{GLIDE}, we compare it with a broad set of representative baselines across the three main benchmarks.
These baselines cover five methodological paradigms, reflecting different reasoning structures, interaction modes, and supervision requirements.

\noindent \textbf{Linear and Standard Prompting.}
This category represents the foundational capabilities of LLMs without explicit search or multi-agent control.

\begin{itemize}
    \item \textit{Standard Prompting (IO) \& Single LLM:}
    These baselines use direct input-output generation with vanilla prompts, serving as basic reference points for evaluating the benefit of structured search and heterogeneous agent control.

    \item \textit{Chain-of-Thought (CoT)~\citep{wei2022chaincot}:}
    CoT prompts the model to generate intermediate reasoning steps before producing the final answer, decomposing complex problems into linear reasoning traces.

    \item \textit{Self-Consistency (CoT-SC)~\citep{wang2022self-con}:}
    CoT-SC samples multiple reasoning paths and aggregates them through majority voting.
    It provides a strong baseline for evaluating the effect of sampling-based consensus without explicit tree search or agent-level calibration.
\end{itemize}

\noindent \textbf{Feedback-Driven and Interactive Reasoning.}
These methods extend linear generation by incorporating interaction, critique, or iterative refinement.

\begin{itemize}
    \item \textit{ReAct~\citep{yao2023react}:}
    ReAct interleaves reasoning and acting, allowing the model to produce thought-action trajectories and interact with external environments.
    This design helps reduce hallucination by grounding reasoning in observations.

    \item \textit{Reflexion~\citep{shinn2023reflexion}:}
    Reflexion uses verbal reinforcement feedback to support iterative self-correction.
    It stores reflections over previous failures and uses them to improve subsequent trials without updating model parameters.
\end{itemize}

\noindent \textbf{Structured Search and Planning.}
These approaches formulate reasoning as state-space search, using tree, graph, beam, or MCTS structures to explore multiple solution paths.

\begin{itemize}
    \item \textit{Tree, Graph, and Beam Search:}
    Methods such as ToT~\citep{tot}, GoT~\citep{got}, MDToC~\citep{ta2025mdtoc}, and AgentKit~\citep{wu2024agentkit} generalize linear reasoning chains into non-linear search structures, enabling multiple candidate branches to be explored and compared.
    For retrieval-intensive reasoning, we also include Beam Retrieval~\citep{zhang2023end}, which applies beam-search-style selection to iteratively rank and select evidence chains.

    \item \textit{MCTS-based Agents:}
    RAP~\citep{hao2023reasoning} and LATS~\citep{zhou2024language} integrate Monte Carlo Tree Search with LLM agents.
    LATS, in particular, unifies reasoning, acting, and planning by using an external LLM-as-a-Judge to estimate state values and guide search toward high-reward trajectories.
\end{itemize}

\noindent \textbf{Multi-Agent Frameworks.}
These systems leverage multiple LLM agents to diversify reasoning trajectories and reduce single-model bias.

\begin{itemize}
\item \textit{MoA~\citep{wang2024mixture_moa}:}
MoA aggregates responses from multiple agents through layered collaboration, aiming to improve answer quality via multi-agent response refinement.

\item \textit{MASTER~\citep{gan2025master}:}
MASTER is a multi-agent MCTS framework that combines agent collaboration with tree-structured search, using multi-agent interactions to guide reasoning and decision making.

    \item \textit{SYMPHONY~\citep{symphony}:}
    SYMPHONY optimizes query dispatching across a heterogeneous model pool.
    We compare with two variants when available: SYMPHONY-L, which uses a high-cost pool including proprietary or high-capacity models, and SYMPHONY-S, which uses open-source models.
    SYMPHONY-S serves as the primary controlled baseline because it uses the same open-weight model pool as \textsc{GLIDE}.
\end{itemize}

\noindent \textbf{Task-Specific Methods for WebShop.}
For WebShop, we additionally include task-native and supervised methods.
These include Imitation Learning (IL), IL+RL, and Human Expert baselines from the original WebShop benchmark~\citep{yao2022webshop}, as well as WebGUM~\citep{furuta2024multimodal}, a multimodal agent fine-tuned on web navigation and instruction-following data.

\vspace{0.5em}
\paragraph{Result Sources and Backbone Fairness.}
To ensure transparent comparison, we follow the reporting protocols and backbone settings of prior work, while using SYMPHONY-S as the primary controlled baseline for \textsc{GLIDE}.

\begin{itemize}
    \item \textit{HotpotQA.}
    Baselines are based on the GPT-4 backbone.
    Except for SYMPHONY, reported results are cited from~\citet{gan2025master}.

    \item \textit{WebShop.}
    Most reported baselines use the GPT-4 backbone.
    SYMPHONY results are cited from the original SYMPHONY paper, while other GPT-4-based baselines are adopted from~\citet{gan2025master}.

    \item \textit{Game of 24.}
    Baselines predominantly use the GPT-3.5-Turbo backbone.
    Single-LLM results are cited from~\citet{llmreasoner}, GoT and MDToC results from~\citet{ta2025mdtoc}, and the remaining baselines from~\citet{zhou2024language}.

    \item\textit{Controlled comparison.}
SYMPHONY-S is reproduced with the same open-weight model pool and comparable search budget as \textsc{GLIDE}, making it the primary controlled baseline across all three main benchmarks.
\end{itemize}

\section{Supplementary Analyses for Main Experiments}
\label{appendix:main_supplementary_analyses}

This section provides additional analyses for the main experiments on HotpotQA, WebShop, and Game of 24.
These analyses further examine the efficiency, parameter behavior, and step-level preference quality of \textsc{GLIDE} under the same MCTS-based heterogeneous agent setting used in the main benchmark results.

\subsection{Test-Time Compute Scaling}
\label{appendix:cost}

A major limitation of tree-structured reasoning frameworks is the inference overhead introduced by expanding and evaluating many candidate nodes.
We evaluate the test-time compute efficiency of \textsc{GLIDE} on HotpotQA by scaling the MCTS budget from \(K=10\) to \(K=50\).
For each setting, we report Exact Match, the average number of expanded nodes, and average token consumption per question.
Following prior reporting protocols, the results of ToT, RAP, and LATS are taken from LATS~\citep{zhou2024language}, and the SYMPHONY results are taken from SYMPHONY~\citep{symphony}.

Many structured reasoning methods follow a generate-then-verify pipeline, where node expansion is coupled with additional verifier or LLM-as-a-Judge calls.
This design increases both token consumption and latency, especially when the search budget grows.
In contrast, \textsc{GLIDE} computes its value signal from hidden states and predictive distributions already produced during generation, avoiding additional verifier-generated tokens.

Table~\ref{tab:token_cost} shows that \textsc{GLIDE} scales favorably with test-time compute.
At \(K=10\), \textsc{GLIDE} substantially improves over SYMPHONY-S while using fewer tokens than SYMPHONY-L.
When the budget is increased to \(K=50\), \textsc{GLIDE} outperforms ToT, RAP, and LATS while using far fewer tokens.
These results indicate that verifier-free intrinsic evaluation and adaptive branching provide a more cost-effective scaling path than external-judge-based search.

\subsection{Algorithmic Parameter Analysis}
\label{appendix:parameter_analysis}

We analyze the main hyperparameters governing \textsc{GLIDE}'s inference behavior.
The analysis focuses on two aspects: the search budget that controls MCTS complexity, and the online distributional horizon used for calibration.

\subsubsection{Search Budget and Adaptive Branching}
\label{sub:search_params}

In MCTS, inference cost is primarily controlled by the number of search iterations and the branching width.
High-resource baselines such as RAP~\citep{hao2023reasoning} and LATS~\citep{zhou2024language} often rely on larger search budgets to improve exploration and reduce value-estimation variance, but this brute-force scaling strategy is costly.
\textsc{GLIDE} instead uses a constrained search budget and allocates branching adaptively according to predictive uncertainty and calibrated reward saturation.

\paragraph{Trajectory budget.}
The maximum number of MCTS iterations is set to \(K=10\) across HotpotQA, WebShop, and Game of 24.
This budget follows the efficient SYMPHONY-S setting and keeps inference cost controlled across all tasks.
Under this constraint, performance improvements must come from more effective candidate prioritization and search control rather than from simply increasing the number of simulations.

\paragraph{Branching width.}
The branching width determines how many candidate continuations are expanded from a search state.
Standard fixed-width methods allocate the same expansion budget to all states, regardless of whether the current state is ambiguous, already promising, or unlikely to benefit from further exploration.
In \textsc{GLIDE}, the branching width \(n(x,a)\) is adapted using Eq.~\eqref{eq:adaptive_branching}.
We set \(n_{\min}=1\) and \(n_{\max}=4\), where \(n_{\max}=4\) matches the fixed expansion width used by SYMPHONY.
Thus, \textsc{GLIDE} never exceeds the maximum branching width of the controlled baseline, while its average branching width can be lower because easy or saturated states receive fewer expansions.
Figure~\ref{fig:ablation_study}(a) further analyzes the effect of different \(n_{\max}\) settings.

\subsubsection{Calibration Horizon and Stability}
\label{sub:calibration_window}

The sliding windows \(\mathcal{W}_i^S\) and \(\mathcal{W}_i^H\) define the online horizon for distributional calibration.
We set both window sizes to 50:
\[
|\mathcal{W}_i^S|=|\mathcal{W}_i^H|=50 .
\]
The score window \(\mathcal{W}_i^S\) is used for the distributional rank score in Eq.~\eqref{eq:rank_score}, while the entropy window \(\mathcal{W}_i^H\) is used for relative uncertainty estimation in Eq.~\eqref{eq:relative_entropy}.

These windows serve as online, agent-specific calibrators rather than static storage buffers.
Unlike offline normalization methods that require pre-computed statistics on a held-out set, \textsc{GLIDE} builds each agent's local score and entropy profiles during inference.
This design is important in multi-step reasoning, where the distribution of generated tokens and intrinsic scores can shift as context length and task difficulty evolve.

The window also provides a forgetting mechanism.
A very small window reacts quickly but can be sensitive to transient outliers, causing unstable rank and entropy estimates.
A very large window provides smoother statistics but may introduce distributional lag, making the calibration slow to adapt to changes in reasoning depth or local task complexity.
Figure~\ref{fig:ablation_study}(b) shows that a window size of 50 provides a stable trade-off between sensitivity and robustness.

\subsection{MCTS Evaluation Baselines under Fixed Search}
\label{app:mcts_eval_baselines}

As an extension of the ablation study in Table~\ref{tab:ablation_wrap}, we evaluate different scoring functions under the same fixed MCTS framework.
This experiment isolates the effect of the evaluation signal while keeping the search procedure unchanged.
The compared baselines include probability-based confidence metrics, hidden-state-based metrics, supervised process reward models, and random heterogeneous-agent evaluation.

\begin{table}[t]
\centering
\caption{
\textbf{MCTS ablation with different evaluation baselines.}
All variants are evaluated under the same fixed MCTS framework.
Hot denotes HotpotQA EM, Web denotes WebShop SR, and G24 denotes Game of 24 success rate.
}
\label{tab:mcts_eval_baselines}
\small
\setlength{\tabcolsep}{5pt}
\begin{adjustbox}{max width=\linewidth}
\begin{tabular}{lccc}
\toprule
\textbf{Variant} & \textbf{Hot} & \textbf{Web} & \textbf{G24} \\
\midrule
Perplexity & 0.61 & 0.37 & 0.18 \\
Energy & 0.54 & 0.38 & 0.53 \\
LN-Entropy & 0.63 & 0.31 & 0.45 \\
CoE-R & 0.61 & 0.47 & 0.20 \\
CoE-C & 0.71 & 0.45 & 0.51 \\
GenPRM-7B & 0.61 & 0.34 & 0.47 \\
Rand. Agent & 0.59 & 0.56 & 0.43 \\
\rowcolor{gray!10}
\textbf{\textsc{GLIDE}} & \textbf{0.73} & \textbf{0.57} & \textbf{0.54} \\
\bottomrule
\end{tabular}
\end{adjustbox}
\end{table}

Table~\ref{tab:mcts_eval_baselines} shows that \textsc{GLIDE} achieves the best performance across all three domains.
Compared with probability-based metrics such as perplexity, energy, and length-normalized entropy, \textsc{GLIDE} provides a more effective search-control signal.
It also outperforms CoE variants and GenPRM-7B, suggesting that the proposed combination of $S_{\mathrm{ISLE}}$ and distributional calibration is better suited for heterogeneous MCTS selection.

\subsection{Step-Level Preference and Efficiency Analysis}
\label{app:step_preference_details}

The step-level preference analysis in Section~\ref{subsec:preference_quality} is derived from the search traces of the main MCTS experiments.
For each decision point, we collect the expanded child nodes and evaluate whether different scoring signals rank successful continuations ahead of unsuccessful ones.
A child node is labeled according to the final environment outcome reached from that node.
This protocol tests whether a scoring method provides a useful preference signal for MCTS selection, rather than whether it certifies the correctness of an isolated reasoning step.

\subsubsection{Scoring Baselines}
\label{app:discriminative_metrics}

We compare $R_{\mathrm{GLIDE}}$ with probability-based intrinsic metrics, hidden-state-based metrics, uncalibrated intrinsic scores, heterogeneous judge scores, and process reward models.
Unless otherwise specified, all baselines are converted to a common higher-is-better preference orientation before ranking child nodes.

\noindent \textbf{Intrinsic uncertainty metrics.}
These baselines are training-free confidence metrics computed from token probabilities or logits during generation.

\begin{itemize}
    \item \textit{Perplexity~\citep{liu2020energy}.}
    Perplexity summarizes sequence-level uncertainty from the likelihood assigned to the generated continuation.
    Since lower perplexity indicates higher model confidence, we use its negative value as the ranking score.

    \item \textit{Length-normalized entropy~\citep{malinin2020uncertainty}.}
    Length-normalized entropy averages the Shannon entropy of the predictive distribution over decoding steps.
    Lower entropy indicates a more concentrated next-token distribution, so we use its negative value as the ranking score.

    \item \textit{Energy score~\citep{huang2023look}.}
    The energy score is computed from unnormalized model logits and aggregated over the generated continuation.
    We choose the sign convention so that larger values correspond to stronger confidence, making it comparable with other preference scores.
\end{itemize}

\noindent \textbf{Hidden-state-based metrics.}
These baselines use internal representations rather than only token probabilities.

\begin{itemize}
    \item \textit{Chain-of-Embedding Metrics (CoE)~\citep{coe}.}
    CoE measures hidden-state trajectories across layers and uses their magnitude and angular changes as indicators of generation reliability.
    We compare with both CoE-R and CoE-C variants following the original implementation.
    CoE-R combines normalized magnitude and angle changes in real space, while CoE-C maps layer-wise changes into a complex-plane representation to capture their joint evolution.

    \item \textit{CoT-Kinetics~\citep{cotkinetics}.}
    CoT-Kinetics measures reasoning reliability by formulating the layer-wise evolution of reasoning-token hidden states as a kinetic process.
    It computes a scalar energy score from semantic momentum, semantic curvature, and output uncertainty, thereby assessing the soundness of the reasoning trajectory using only internal model states.
    We do not include CoT-Kinetics in the MCTS-based experiments, since it is primarily designed for long chain-of-thought reasoning in LRMs, whereas our method focuses on step-level decision control.
    
\end{itemize}

\noindent \textbf{Heterogeneous and reward-model baselines.}
These baselines evaluate whether performance gains come from agent heterogeneity, calibrated intrinsic scoring, or supervised process supervision.

\begin{itemize}
    \item \textit{Raw \(S_{\mathrm{ISLE}}\).}
    This baseline directly uses the uncalibrated intrinsic stability score from each generating agent.
    It tests whether agent-wise distributional calibration is necessary.

    \item \textit{Rank-only score.}
    This baseline uses only \(S_{\mathrm{rank}}^{(i)}\) from Eq.~\eqref{eq:rank_score}, discarding the absolute intrinsic score.
    It evaluates whether relative standing alone is sufficient for robust search control.

    \item \textit{Hetero-Raw.}
    This control uses heterogeneous agent outputs without calibrated intrinsic reward.
    It tests whether gains arise merely from access to multiple models rather than from \textsc{GLIDE}'s distributional calibration.

    \item \textit{GenPRM-7B~\citep{zhao2025genprm}.}
    GenPRM-7B is a process reward model designed to provide granular step-level feedback.

    \item \textit{PRM-RLHFlow~\citep{rlhflow}.}
    We use the Llama3.1-8B-PRM-Deepseek-Data variant, which leverages preference data for process-level verification.

    \item \textit{PRM-Skywork~\citep{skywork}.}
    We use the Skywork-o1-Open-PRM-Qwen-2.5-1.5B variant, which provides a lightweight process reward model for step evaluation.
\end{itemize}

Unlike PRM-based baselines, \textsc{GLIDE} does not require task-specific process supervision or additional verifier forward passes.
Its calibrated reward \(R_{\mathrm{GLIDE}}\) is computed from the same generation pass used to produce the candidate continuation.

\subsubsection{Ranking Metrics}
\label{app:step_preference_metrics}

We report P@1, MRR, and NDCG@3 for step-level preference evaluation.
Let \(\mathcal{Q}\) denote the set of evaluated selection points.
For a selection point \(q\in\mathcal{Q}\), let \(\pi_q(k)\) be the child node ranked at position \(k\) by a scoring method, and let \(g_q(v)\in\{0,1\}\) indicate whether child node \(v\) eventually reaches a successful final outcome.

Precision at rank 1 is
\begin{equation}
\mathrm{P@1}
=
\frac{1}{|\mathcal{Q}|}
\sum_{q\in\mathcal{Q}}
g_q(\pi_q(1)).
\end{equation}
MRR measures how early the first successful child appears:
\begin{equation}
\mathrm{MRR}
=
\frac{1}{|\mathcal{Q}|}
\sum_{q\in\mathcal{Q}}
\frac{1}{r_q}.
\end{equation}

\begin{equation}
r_q
=
\min\{k:\,g_q(\pi_q(k))=1\}.
\end{equation}
If no successful child exists for a selection point, the reciprocal-rank contribution is set to zero.
NDCG@3 measures the quality of the top-ranked child nodes:
\begin{equation}
\mathrm{NDCG@3}
=
\frac{1}{|\mathcal{Q}|}
\sum_{q\in\mathcal{Q}}
\frac{\mathrm{DCG@3}_q}{\mathrm{IDCG@3}_q},
\end{equation}
where
\begin{equation}
\mathrm{DCG@3}_q
=
\sum_{k=1}^{3}
\frac{
g_q(\pi_q(k))
}{
\log_2(k+1)
}.
\end{equation}
We also report inference latency and peak GPU memory to quantify the efficiency of each scoring method.

\subsection{Failure Case Analysis}
\label{app:failure_case_analysis}

We further analyze cases where \textsc{GLIDE} assigns high scores to incorrect steps.
The most common failure pattern is a \emph{confident semantic mismatch}: the generated step is internally coherent and lexically well supported by the current context, but it fails to satisfy an external task constraint.
In such cases, the model's layer-wise residual evolution can remain stable, causing $S_{\mathrm{ISLE}}$ to assign a high intrinsic score even though the step is ultimately wrong.

Table~\ref{tab:failure_case_webshop} shows a representative example from WebShop.
Given the instruction ``vegetarian smoked peppered bacon,'' the agent clicked a product titled ``Smoked Bacon Sea Salt 3-Pack.''
The step matches salient surface cues such as ``smoked,'' ``bacon,'' and related product descriptors, leading to a confident and coherent internal trajectory.
However, the clicked item is a seasoning product rather than a vegetarian bacon substitute, making the action semantically incorrect.

\begin{table}[t]
\centering
\caption{
\textbf{Representative failure case.}
\textsc{GLIDE} can overscore internally coherent but externally mismatched steps.
}
\label{tab:failure_case_webshop}
\small
\setlength{\tabcolsep}{4pt}
\begin{adjustbox}{max width=\linewidth}
\begin{tabular}{p{0.25\linewidth} p{0.31\linewidth} p{0.34\linewidth}}
\toprule
\textbf{Instruction} & \textbf{High-scoring step} & \textbf{Failure mode} \\
\midrule
Vegetarian smoked peppered bacon
&
Click ``Smoked Bacon Sea Salt 3-Pack''
&
Strong lexical overlap makes the step internally coherent, but the product is not a vegetarian bacon substitute.
\\
\bottomrule
\end{tabular}
\end{adjustbox}
\end{table}

This failure mode highlights a natural boundary of intrinsic evaluation.
$S_{\mathrm{ISLE}}$ measures whether the model's internal computation evolves coherently within the current generation step; it does not directly verify external semantic grounding or task-specific constraints.
Therefore, high intrinsic coherence should be interpreted as a preference signal for search control rather than a certificate of correctness.

In practice, the MCTS framework can partially mitigate such errors.
If later observations expose the mismatch, subsequent node evaluations may decrease and alternative branches can still be explored.
Nevertheless, recovery is not guaranteed when the incorrect branch remains superficially plausible or the search budget is limited.
This suggests a promising direction for future work: combining calibrated intrinsic evidence with lightweight grounding checks or environment-aware constraint signals, while preserving the low-overhead advantages of \textsc{GLIDE}.


\section{Cross-Task Outcome Separability}
\label{appendix:outcome_separability_details}

This section provides details for the auxiliary cross-task outcome-separability analysis in Section~\ref{subsec:outcome_separability}.
Unlike the main search experiments, this evaluation does not use search traces, tree expansion, or adaptive branching.
Each model directly generates one solution for each input, and the solution is labeled by its final task outcome.
The goal is to test whether $S_{\mathrm{ISLE}}$ remains associated with final outcomes under direct generation, outside the main MCTS control pipeline.

\subsection{Datasets and Models}
\label{app:outcome_model_config}

We evaluate outcome separability across three complementary settings: multilingual mathematical reasoning, scientific knowledge reasoning, and code generation.

\begin{description}
    \item[Multilingual mathematical reasoning.]
    MGSM~\citep{data:mgsm} is a subset of GSM8K with 11 language versions.
    We evaluate all languages: English, Bengali, German, Spanish, French, Japanese, Russian, Swahili, Telugu, Thai, and Chinese.
    Unless otherwise specified, MGSM results report the average over these 11 languages.
    Averaging across languages helps reduce dependence on language-specific surface forms, testing whether the scoring signal reflects model-internal outcome evidence rather than superficial linguistic variation.

    \item[Scientific knowledge reasoning.]
    GPQA Diamond~\citep{data:gpqa} is a PhD-level multiple-choice question answering benchmark covering physics, chemistry, and biology.
    It evaluates outcome separability under specialized domain knowledge and difficult scientific reasoning.

    \item[Code generation.]
    HumanEval~\citep{data:humaneval} consists of programming problems paired with unit tests.
    A generated solution is labeled successful if it passes the corresponding tests, allowing us to evaluate outcome separability for executable program synthesis.
\end{description}

We use open-source LLMs spanning different model families and parameter scales:
Mistral-7B-Instruct-v0.3~\citep{jiang2023mistral7b},
Gemma-3-12B-IT~\citep{model:gemma},
Phi-3-medium-128k-instruct~\citep{model:phi},
InternLM2-Chat-20B~\citep{model:intern},
and Qwen2.5-32B-Instruct~\citep{model:qwen}.
This model set allows us to test whether the scoring signal remains outcome-relevant across architectures, training recipes, and model sizes.

\subsection{Generation Protocol}
\label{app:outcome_generation_protocol}

We use deterministic greedy decoding with \texttt{do\_sample=False}.
The maximum number of new tokens is set to 512 for MGSM and GPQA, and 1024 for HumanEval.
This protocol is designed to test outcome separability under a bounded single-pass generation setting, rather than to improve accuracy through additional test-time scaling such as repeated sampling, self-consistency, or search.

This design is consistent with the role of $S_{\mathrm{ISLE}}$ in the main method.
\textsc{GLIDE} uses $S_{\mathrm{ISLE}}$ as a step-level preference signal during sequential search, where the goal is to rank candidate continuations rather than to elicit long free-form chains.
Accordingly, this auxiliary experiment uses direct generations with moderate token limits, allowing the model to produce a complete answer while keeping the evaluation focused on the intrinsic score's outcome relevance.
The larger budget for HumanEval accounts for full Python implementations, whereas MGSM and GPQA require only concise reasoning and final-answer extraction.

The task-specific prompts are shown in Figure~\ref{fig:outcome_prompt_templates}.
They standardize the final-answer format for MGSM and GPQA and enforce executable Python output for HumanEval, enabling consistent outcome labeling across tasks.

\subsection{Outcome-Separability Results}
\label{appendix:outcome_separability}

We compare $S_{\mathrm{ISLE}}$ with the scoring baselines described in Appendix~\ref{app:discriminative_metrics}, including probability-based confidence metrics, hidden-state-based metrics, and output-based confidence baselines when applicable.
Unlike the step-level analysis in Section~\ref{subsec:preference_quality}, the evaluation unit here is a complete generated solution rather than an expanded child node in the search tree.

For each model--dataset pair, each generated solution is labeled as successful or failed according to the final task outcome.
A scoring method is effective if it assigns higher scores to successful generations than to failed ones.
We report AUROC, AUPR, and FPR95.
AUROC measures the probability that a randomly selected successful generation receives a higher score than a randomly selected failed generation.
AUPR summarizes the precision--recall trade-off and is especially informative when the numbers of successful and failed samples are imbalanced.
FPR95 reports the false-positive rate when the true-positive recall is fixed at 95\%, measuring how many failed generations are incorrectly accepted when most successful generations are retained.
Thus, higher AUROC and AUPR, and lower FPR95, indicate stronger outcome separability.

\begin{figure*}[p]
\centering
\includegraphics[width=\textwidth]{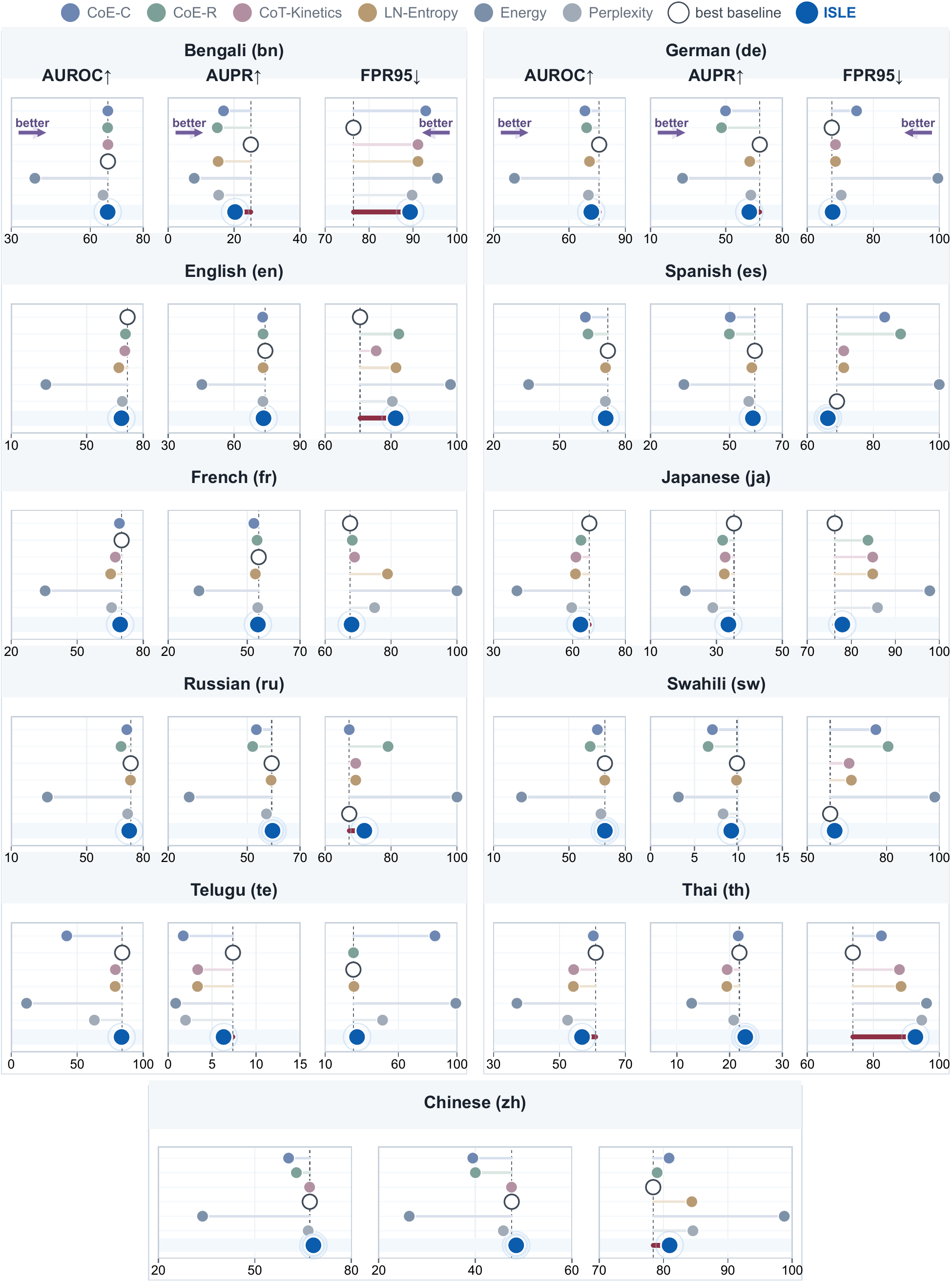}
\caption{
\textbf{Language-level MGSM outcome separability for Mistral-7B-Instruct-v0.3.}
The figure shows the performance of different scoring methods across the 11 MGSM language versions.
Higher AUROC/AUPR and lower FPR95 indicate better separation between successful and failed generations.
}
\label{fig:mgsm-adaptive-plate-mistral}
\end{figure*}

\begin{figure*}[p]
\centering
\includegraphics[width=\textwidth]{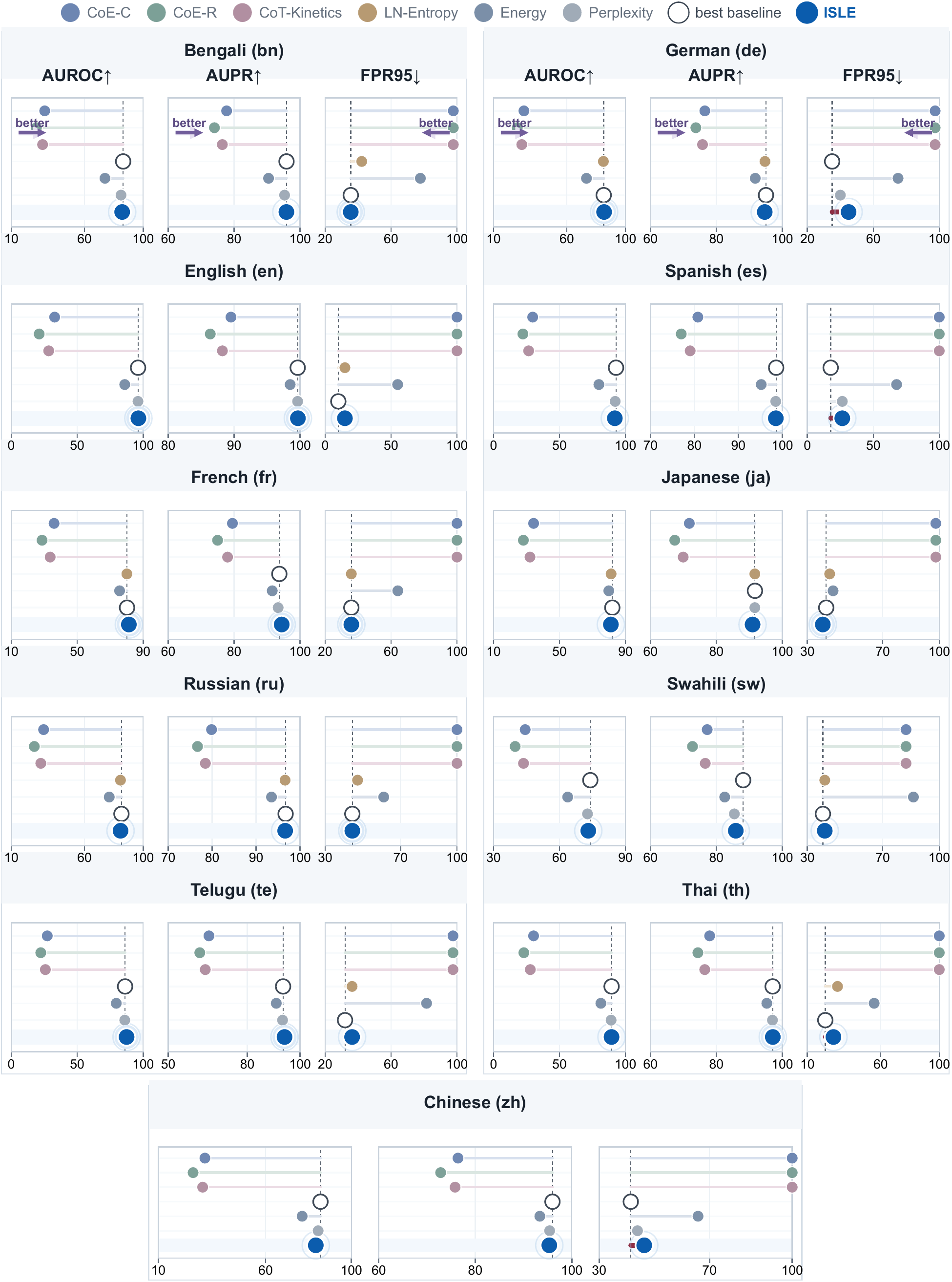}
\caption{
\textbf{Language-level MGSM outcome separability for Gemma-3-12B-IT.}
The figure shows the performance of different scoring methods across the 11 MGSM language versions.
Higher AUROC/AUPR and lower FPR95 indicate better separation between successful and failed generations.
}
\label{fig:mgsm-adaptive-plate-gemma-3-12b}
\end{figure*}

\begin{figure*}[p]
\centering
\includegraphics[width=\textwidth]{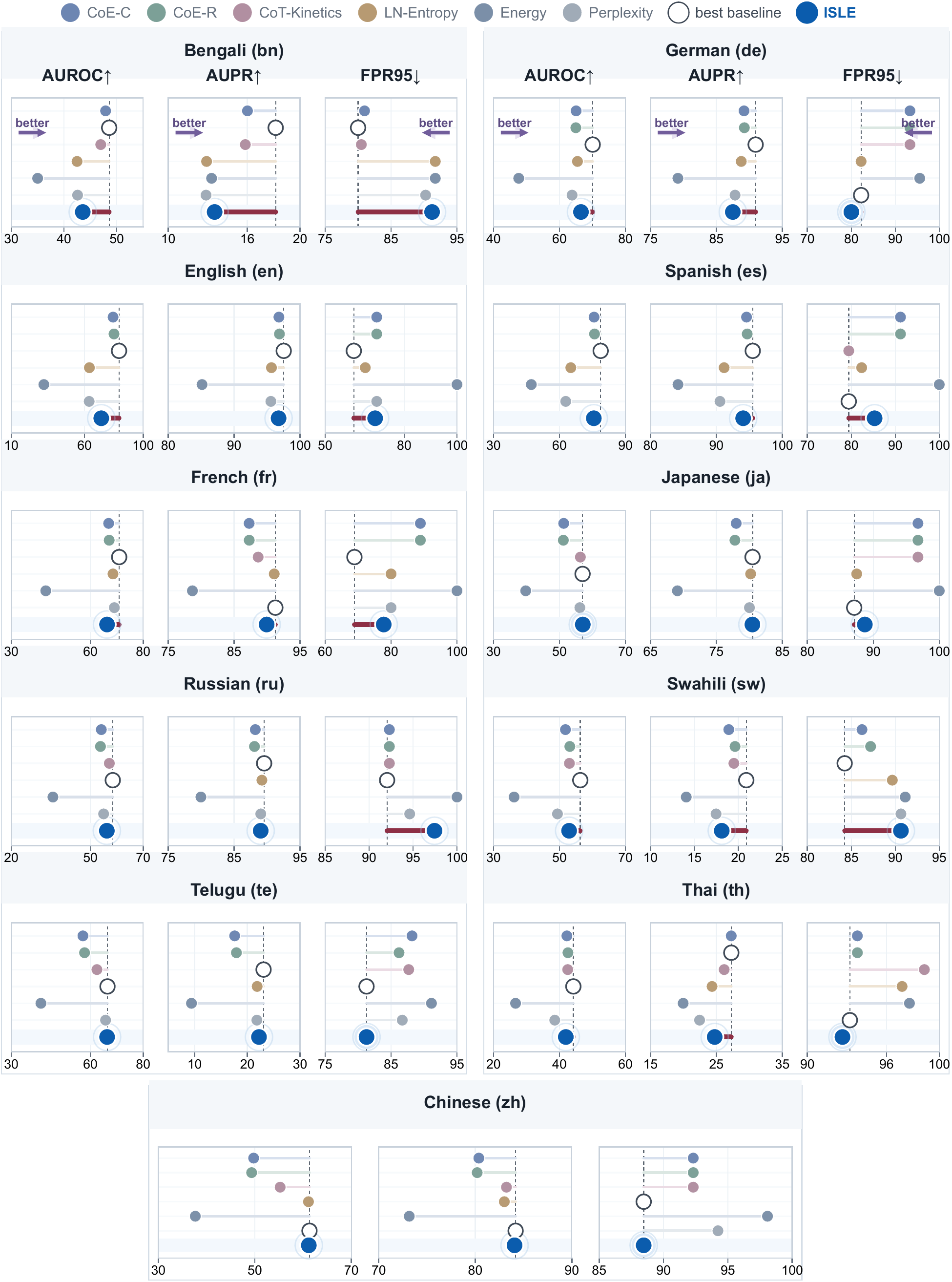}
\caption{
\textbf{Language-level MGSM outcome separability for Phi-3-medium-128k-instruct.}
The figure shows the performance of different scoring methods across the 11 MGSM language versions.
Higher AUROC/AUPR and lower FPR95 indicate better separation between successful and failed generations.
}
\label{fig:mgsm-adaptive-plate-Phi-14B}
\end{figure*}

\begin{figure*}[p]
\centering
\includegraphics[width=\textwidth]{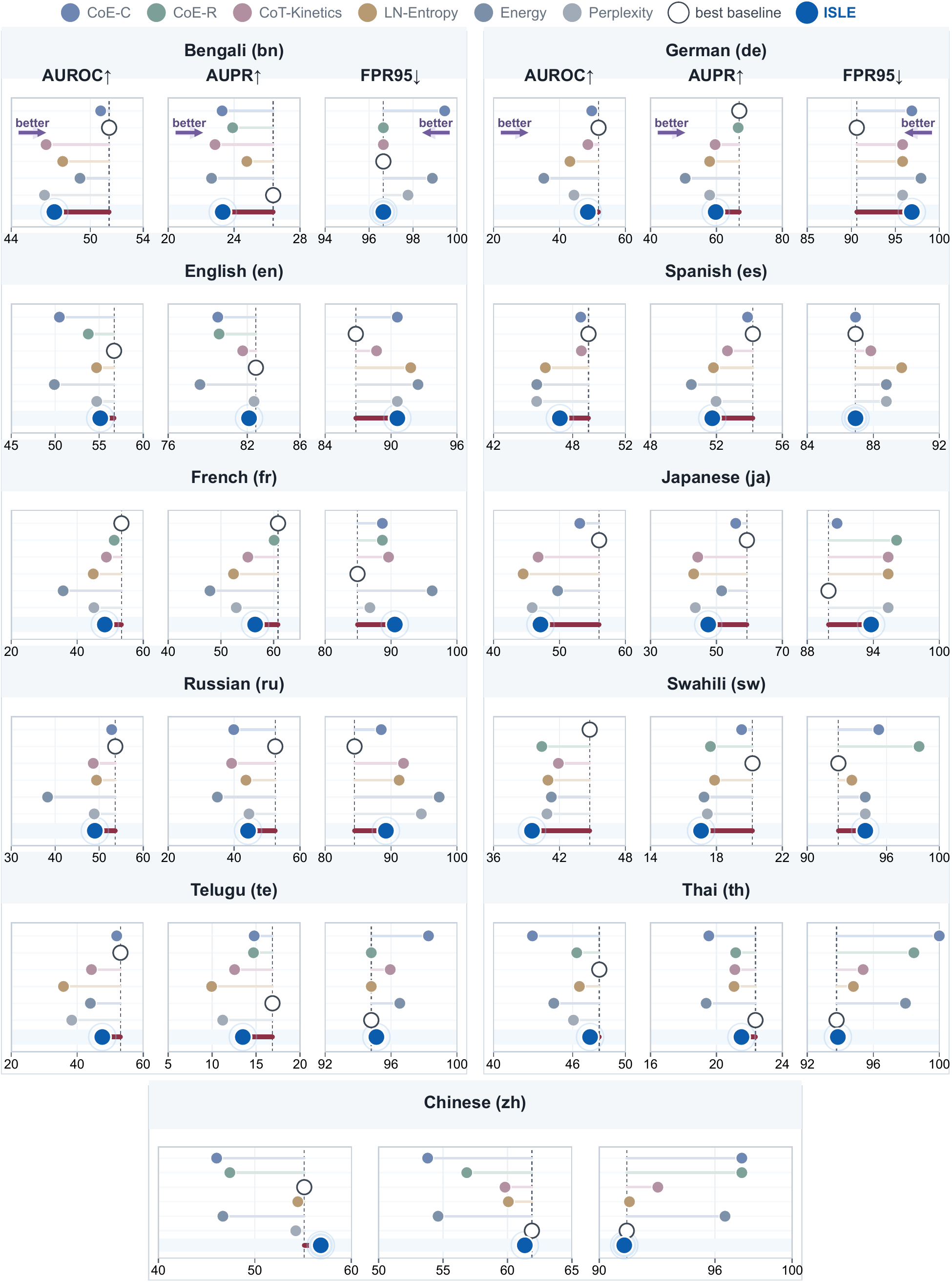}
\caption{
\textbf{Language-level MGSM outcome separability for InternLM2-Chat-20B.}
The figure shows the performance of different scoring methods across the 11 MGSM language versions.
Higher AUROC/AUPR and lower FPR95 indicate better separation between successful and failed generations.
}
\label{fig:mgsm-adaptive-plate-internlm2}
\end{figure*}

\begin{figure*}[p]
\centering
\includegraphics[width=\textwidth]{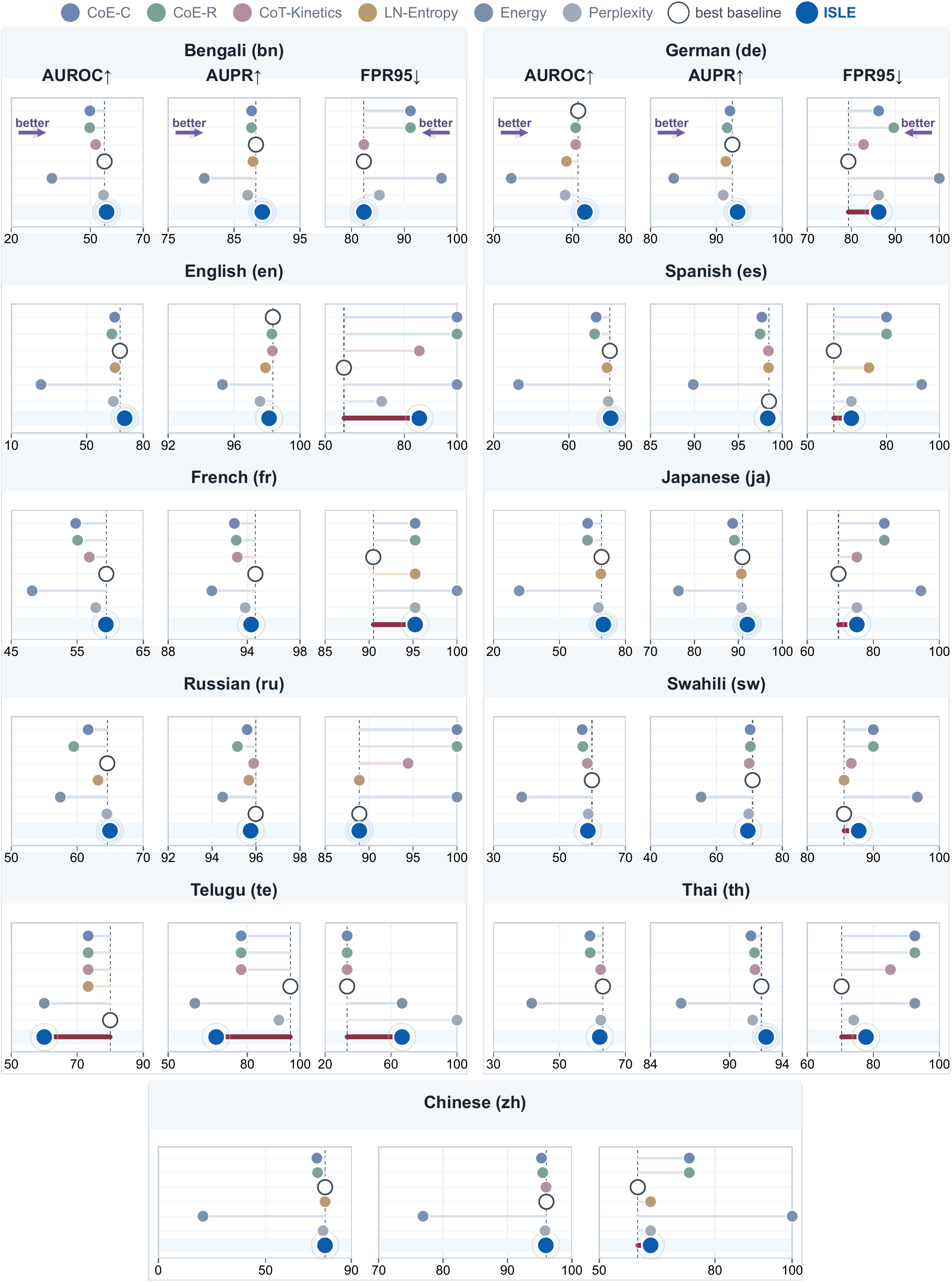}
\caption{
\textbf{Language-level MGSM outcome separability for Qwen2.5-32B-Instruct.}
The figure shows the performance of different scoring methods across the 11 MGSM language versions.
Higher AUROC/AUPR and lower FPR95 indicate better separation between successful and failed generations.
}
\label{fig:mgsm-adaptive-plate-qwen25-32B}
\end{figure*}

Full model-wise and dataset-wise results are reported in Table~\ref{tab:outcome_separability_full}.
For MGSM, the table reports the average over 11 language versions, while language-level results for each model are visualized in Figures~\ref{fig:mgsm-adaptive-plate-mistral}--\ref{fig:mgsm-adaptive-plate-qwen25-32B}.
Across datasets and model families, $S_{\mathrm{ISLE}}$ provides consistent separation between successful and failed generations, supporting the outcome relevance of the intrinsic signal outside the main MCTS control pipeline.

\definecolor{bestBg}{HTML}{DBEAFE}        
\definecolor{secondBg}{HTML}{FED7D7}      
\definecolor{avgBestBg}{HTML}{FEF3C7}     
\definecolor{avgSecondBg}{HTML}{E9D5FF}   

\newcommand{\bestnum}[1]{%
  \begingroup\setlength{\fboxsep}{1.15pt}\colorbox{bestBg}{\strut\textbf{#1}}\endgroup%
}
\newcommand{\secondnum}[1]{%
  \begingroup\setlength{\fboxsep}{1.15pt}\colorbox{secondBg}{\strut #1}\endgroup%
}
\newcommand{\avgbestnum}[1]{%
  \begingroup\setlength{\fboxsep}{1.15pt}\colorbox{avgBestBg}{\strut\textbf{#1}}\endgroup%
}
\newcommand{\avgsecondnum}[1]{%
  \begingroup\setlength{\fboxsep}{1.15pt}\colorbox{avgSecondBg}{\strut #1}\endgroup%
}

\begin{table*}[t]
\centering
\caption{
\textbf{Cross-task outcome separability across models and datasets.}
Each cell reports AUROC/AUPR/FPR95, with higher AUROC/AUPR and lower FPR95 indicating stronger separability.
MGSM results are averaged over 11 language versions; language-level results for each model are shown in Figures~\ref{fig:mgsm-adaptive-plate-mistral}--\ref{fig:mgsm-adaptive-plate-qwen25-32B}.
Metric-wise best and second-best values are highlighted per model--dataset column using
\colorbox{bestBg}{Best} and \colorbox{secondBg}{Second Best};
the Average column uses
\colorbox{avgBestBg}{Best} and \colorbox{avgSecondBg}{Second Best}.
}
\label{tab:outcome_separability_full}
\scriptsize
\setlength{\tabcolsep}{2.8pt}
\renewcommand{\arraystretch}{1.12}
\begin{adjustbox}{max width=\textwidth}
\begin{tabular}{lcccccc}
\toprule
\textbf{Method}
& \makecell{\textbf{Mistral-7B}\\\textbf{Inst. v0.3}}
& \makecell{\textbf{Gemma-3}\\\textbf{12B IT}}
& \makecell{\textbf{Phi-3}\\\textbf{Medium 128k}}
& \makecell{\textbf{InternLM2}\\\textbf{Chat 20B}}
& \makecell{\textbf{Qwen2.5}\\\textbf{32B Inst.}}
& \textbf{Average} \\
\midrule

\multicolumn{7}{c}{\textbf{MGSM} \quad (AUROC$\uparrow$/AUPR$\uparrow$/FPR95$\downarrow$)} \\
\midrule
Perplexity               & 65.55/38.72/74.98 & 85.30/94.67/\bestnum{33.53} & 57.06/63.00/86.18 & 46.38/42.98/93.10 & \secondnum{65.93}/\secondnum{91.28}/81.06 & 64.04/66.13/73.77 \\
Energy                   & 31.37/21.48/98.49 & 77.05/92.26/66.68 & 38.41/55.10/96.85 & 43.62/40.42/95.17 & 39.03/80.91/94.61 & 45.90/58.03/90.36 \\
LN-Entropy               & 67.66/39.46/74.03 & \bestnum{85.61}/\bestnum{95.05}/\secondnum{34.76} & 58.96/63.55/\secondnum{85.24} & 46.20/42.29/92.73 & 65.80/\bestnum{92.03}/\bestnum{72.57} & \avgsecondnum{64.85}/\avgsecondnum{66.48}/\avgbestnum{71.87} \\
CoE-R~\citep{coe}        & 67.31/36.28/73.48 & 26.42/73.83/97.58 & 58.59/63.37/88.32 & \bestnum{50.39}/\bestnum{46.07}/\secondnum{92.67} & 62.65/89.80/84.42 & 53.07/61.87/87.29 \\
CoE-C~\citep{coe}        & 63.95/36.46/77.93 & 32.93/77.51/97.58 & 58.31/63.11/88.50 & \secondnum{49.52}/\secondnum{44.37}/93.89 & 63.08/89.81/84.11 & 53.56/62.25/88.40 \\
CoT-Kinetics             & \secondnum{68.76}/\bestnum{41.20}/\secondnum{71.87} & 30.86/76.26/97.58 & \bestnum{61.54}/\bestnum{64.59}/\bestnum{85.02} & 48.57/42.65/92.96 & 65.92/90.28/\secondnum{75.99} & 55.13/63.00/84.68 \\
\textbf{ISLE}            & \bestnum{69.17}/\secondnum{40.83}/\bestnum{71.66} & \secondnum{85.45}/\secondnum{94.83}/35.48 & \secondnum{59.92}/\secondnum{63.67}/85.68 & 48.53/43.51/\bestnum{92.66} & \bestnum{65.99}/89.77/79.60 & \avgbestnum{65.81}/\avgbestnum{66.52}/\avgsecondnum{73.02} \\
\midrule

\multicolumn{7}{c}{\textbf{GPQA Diamond} \quad (AUROC$\uparrow$/AUPR$\uparrow$/FPR95$\downarrow$)} \\
\midrule
Perplexity               & 52.10/12.40/\secondnum{89.58} & 53.15/20.29/96.86 & 62.24/\secondnum{78.88}/\bestnum{94.44} & 52.46/64.77/\secondnum{90.41} & \bestnum{76.73}/\secondnum{89.66}/\bestnum{73.91} & 59.34/53.20/89.04 \\
Energy                   & 40.24/7.92/95.59 & 61.84/\secondnum{30.67}/76.10 & 55.31/77.02/\secondnum{97.22} & 44.44/60.37/100.00 & 33.25/64.47/100.00 & 47.02/48.09/93.78 \\
LN-Entropy               & 53.36/12.17/91.32 & 51.80/19.55/97.48 & \secondnum{62.58}/78.50/\bestnum{94.44} & 54.07/66.05/\secondnum{90.41} & \secondnum{76.57}/\bestnum{89.86}/82.61 & 59.68/53.23/91.25 \\
CoE-R~\citep{coe}        & \bestnum{62.16}/\secondnum{13.91}/95.04 & 67.68/29.15/\secondnum{73.58} & 55.86/78.19/\secondnum{97.22} & 55.89/67.59/97.26 & 60.87/76.74/82.61 & 60.49/53.12/89.14 \\
CoE-C~\citep{coe}        & \secondnum{60.60}/\bestnum{14.79}/90.51 & \bestnum{69.78}/\bestnum{31.45}/\bestnum{67.92} & 57.72/\secondnum{78.88}/\secondnum{97.22} & \bestnum{64.21}/\bestnum{74.69}/\secondnum{90.41} & 58.70/75.34/82.61 & 62.20/55.03/\avgsecondnum{85.73} \\
CoT-Kinetics             & 53.64/13.24/91.32 & \secondnum{69.15}/30.54/75.47 & \bestnum{66.97}/\bestnum{85.43}/\bestnum{94.44} & 60.03/72.60/\bestnum{87.67} & 75.28/87.27/\secondnum{78.26} & \avgbestnum{65.01}/\avgbestnum{57.82}/\avgbestnum{85.43} \\
\textbf{ISLE}            & 54.46/12.68/\bestnum{87.65} & 65.30/29.32/78.74 & 62.52/78.73/\bestnum{94.44} & \secondnum{61.12}/\secondnum{73.45}/91.78 & 74.32/88.91/91.30 & \avgsecondnum{63.54}/\avgsecondnum{56.62}/88.78 \\
\midrule

\multicolumn{7}{c}{\textbf{HumanEval} \quad (AUROC$\uparrow$/AUPR$\uparrow$/FPR95$\downarrow$)} \\
\midrule
Perplexity               & 66.75/48.33/87.78 & 47.64/70.54/\bestnum{80.56} & \secondnum{54.81}/18.45/\secondnum{88.72} & 57.57/65.13/\bestnum{89.86} & \secondnum{56.01}/\secondnum{78.53}/93.02 & \avgsecondnum{56.56}/56.20/\avgsecondnum{87.99} \\
Energy                   & 32.63/19.47/94.87 & \secondnum{47.82}/\secondnum{74.81}/88.89 & 51.90/\secondnum{19.30}/89.47 & 54.92/58.10/\bestnum{89.86} & 45.94/69.14/\secondnum{88.37} & 46.64/48.16/90.29 \\
LN-Entropy               & 67.68/\secondnum{48.96}/81.79 & 47.04/\bestnum{77.92}/\secondnum{86.11} & 50.46/16.04/90.23 & 57.60/64.60/\secondnum{91.30} & 54.85/76.45/90.70 & 55.53/\avgsecondnum{56.79}/88.03 \\
CoE-R~\citep{coe}        & 61.64/39.70/88.89 & 38.02/70.57/97.22 & 53.30/17.80/100.00 & 53.34/62.24/97.10 & 45.08/70.86/93.02 & 50.28/52.23/95.25 \\
CoE-C~\citep{coe}        & 67.31/45.02/94.87 & 38.22/70.80/97.22 & 53.14/17.81/99.25 & 55.67/65.48/95.65 & 46.46/71.64/90.70 & 52.16/54.15/95.54 \\
CoT-Kinetics             & \bestnum{77.74}/48.88/\bestnum{71.79} & 39.02/71.17/97.22 & 51.10/17.48/95.49 & \secondnum{60.16}/\secondnum{66.87}/92.75 & 47.33/73.21/\secondnum{88.37} & 55.07/55.52/89.12 \\
\textbf{ISLE}            & \secondnum{75.80}/\bestnum{49.84}/\secondnum{79.49} & \bestnum{47.87}/74.33/88.89 & \bestnum{57.41}/\bestnum{20.28}/\bestnum{87.22} & \bestnum{61.99}/\bestnum{69.44}/95.65 & \bestnum{57.28}/\bestnum{79.84}/\bestnum{83.72} & \avgbestnum{60.07}/\avgbestnum{58.75}/\avgbestnum{86.99} \\
\bottomrule
\end{tabular}
\end{adjustbox}
\vspace{0.2em}
\footnotesize{
MGSM reports the average over 11 language versions: en, bn, de, es, fr, ja, ru, sw, te, th, and zh.
The ``Average'' column reports the arithmetic mean over the five listed models.
}
\end{table*}

\subsection{Layer-Subset Analysis}
\label{app:layer_subset_analysis}

The default computation of $S_{\mathrm{ISLE}}$ uses all Transformer layers.
This choice follows directly from the theoretical formulation in Section~\ref{sec:theoretical_analysis}, where the score compares local residual updates with the global step update
$\mathbf{u}_{\mathrm{global}}=\mathbf{h}^{(L)}-\mathbf{h}^{(0)}$ accumulated across the full network depth.
Using only a subset of layers weakens this cumulative-trajectory interpretation and introduces a large, hard-to-interpret combinatorial design space.

To assess whether partial trajectories still retain useful outcome-related evidence, we conduct an additional layer-subset analysis on MGSM.
Building on the same outcome-separability protocol, we evaluate four deterministic layer subsets: First 50\%, Last 50\%, Odd, and Even layers.
For each subset, $S_{\mathrm{ISLE}}$ is recomputed using only the selected layer transitions, while the generation outputs and success/failure labels remain unchanged.
This isolates the effect of layer coverage from generation quality.

\begin{figure*}[p]
\centering
\includegraphics[width=\textwidth]{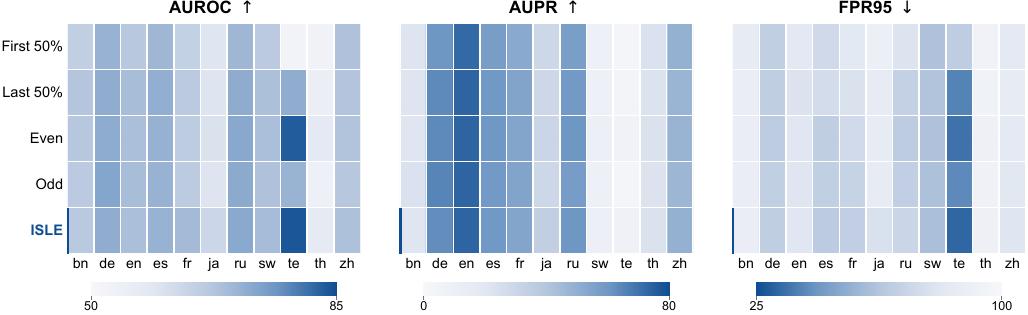}
\caption{
\textbf{Language-level MGSM layer-subset analysis for Mistral-7B-Instruct-v0.3.}
}
\label{fig:mgsm-layer-ablation-mistral}
\end{figure*}

\begin{figure*}[p]
\centering
\includegraphics[width=\textwidth]{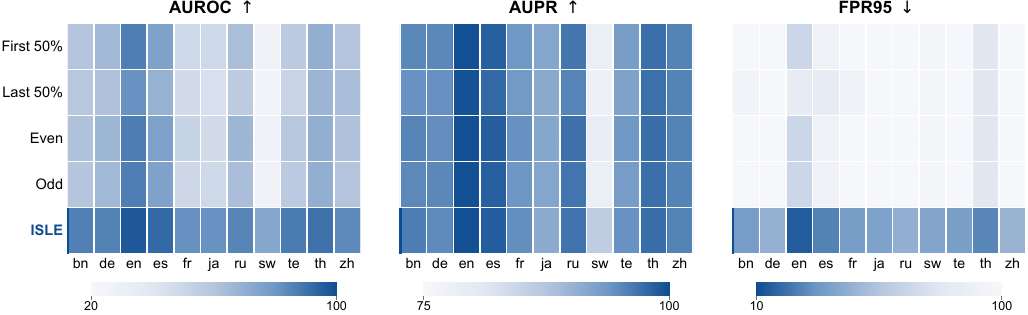}
\caption{
\textbf{Language-level MGSM layer-subset analysis for Gemma-3-12B-IT.}
}
\label{fig:mgsm-layer-ablation-gemma-3-12b}
\end{figure*}

\begin{figure*}[p]
\centering
\includegraphics[width=\textwidth]{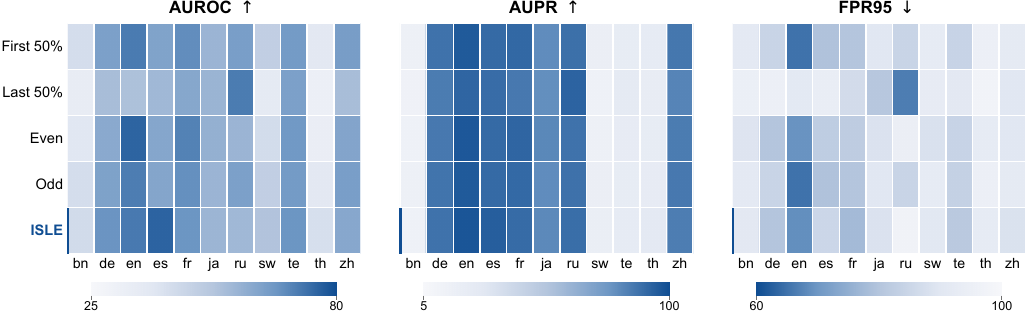}
\caption{
\textbf{Language-level MGSM layer-subset analysis for Phi-3-medium-128k-instruct.}
}
\label{fig:mgsm-layer-ablation-phi3}
\end{figure*}

\begin{figure*}[p]
\centering
\includegraphics[width=\textwidth]{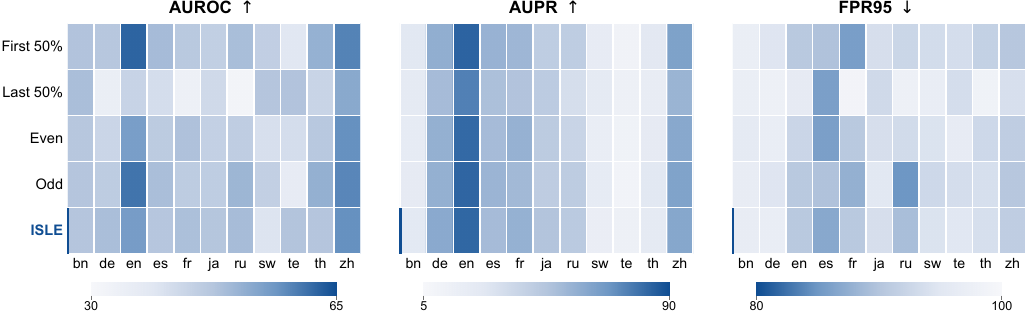}
\caption{
\textbf{Language-level MGSM layer-subset analysis for InternLM2-Chat-20B.}
}
\label{fig:mgsm-layer-ablation-internlm2}
\end{figure*}

\begin{figure*}[p]
\centering
\includegraphics[width=\textwidth]{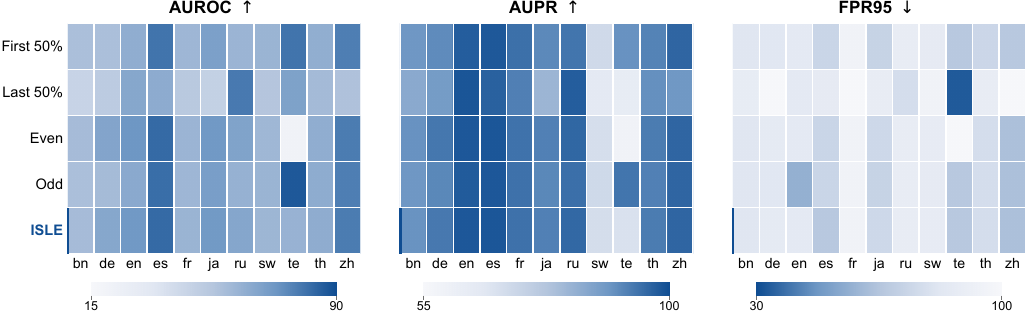}
\caption{
\textbf{Language-level MGSM layer-subset analysis for Qwen2.5-32B-Instruct.}
}
\label{fig:mgsm-layer-ablation-qwen25-32b}
\end{figure*}

The results show that partial layer subsets can still capture useful residual-coherence signals, remaining competitive with several scoring baselines in the full outcome-separability analysis.
However, the all-layer configuration remains the default because it best matches the cumulative residual-stream assumption underlying $S_{\mathrm{ISLE}}$.
Language-level MGSM results for the layer-subset analysis are visualized for each model in Figures~\ref{fig:mgsm-layer-ablation-mistral}--\ref{fig:mgsm-layer-ablation-qwen25-32b}.

\subsection{Memory Overhead of Hidden-State Scoring}
\label{app:vram_overhead}

A potential concern is that computing layer-wise hidden-state dynamics may require storing high-dimensional hidden states across layers, candidate branches, or search trajectories, thereby introducing substantial memory overhead.
In \textsc{GLIDE}, this is not the case.
The intrinsic score is computed on the fly during the generation forward pass: hidden states are extracted for the current candidate, pooled across generated tokens, used to compute the layer-wise residual updates and the corresponding $S_{\mathrm{ISLE}}$ score, and then immediately discarded.
The search tree stores only scalar quantities such as $S_{\mathrm{ISLE}}$ and $R_{\mathrm{GLIDE}}$, rather than full hidden-state trajectories.

To empirically verify this, we measure the peak VRAM increase during the expansion phase on GPQA Diamond.
For each model, we compare peak memory usage with and without \textsc{GLIDE}'s hidden-state scoring under the same generation setting.
As shown in Table~\ref{tab:vram_overhead}, the additional memory footprint is negligible, even when scaling from an 8B model to a 70B model.
This confirms that \textsc{GLIDE}'s intrinsic evaluation introduces little memory overhead in practice.

\begin{table}[t]
\centering
\caption{
\textbf{Peak VRAM overhead of \textsc{GLIDE} hidden-state scoring.}
We report the relative peak memory increase during expansion on GPQA Diamond.
}
\label{tab:vram_overhead}
\small
\setlength{\tabcolsep}{8pt}
\begin{adjustbox}{max width=\linewidth}
\begin{tabular}{lc}
\toprule
\textbf{Model} & \textbf{Peak Memory Increase (\%)} \\
\midrule
Llama-3.1-8B-Instruct  & +1.06 \\
Llama-3.3-70B-Instruct & +2.39 \\
\bottomrule
\end{tabular}
\end{adjustbox}
\end{table}

\section{Use of AI Assistants}
\label{useai}
We utilized AI assistants to help with language editing and writing refinement. All technical content, experimental results, and scientific claims were verified by the authors.

\section{Artifacts Statements}
\label{app:artifacts}

\subsection{Model Artifacts}
All model artifacts are publicly available and used from official open-source releases or public repositories in compliance with their licenses, model cards, and usage terms. We use them only for research evaluation and analysis, without redistributing modified checkpoints; configuration details are provided in Section~\ref{app:outcome_model_config}, \ref{app:discriminative_metrics}, and \ref{subsec:settings}.

\subsection{Data Artifacts}
We employ publicly available benchmarks and task environments for evaluation, including HotpotQA, WebShop, Game of 24, MGSM, GPQA, and HumanEval. These benchmarks are widely used in the open research community. We use them strictly for non-commercial research purposes and comply with their respective licenses, terms of use, and citation requirements.


\begin{figure*}[t]
\centering
\caption{
\textbf{Prompt templates for cross-task outcome separability.}
We use task-specific prompts to standardize answer extraction across mathematical reasoning, scientific multiple-choice reasoning, and code generation.
}
\label{fig:outcome_prompt_templates}
\vspace{0.3em}

\begin{tcolorbox}[
    enhanced,
    width=0.96\textwidth,
    colback=blue!2,
    colframe=blue!45!black,
    boxrule=0.55pt,
    arc=2pt,
    left=6pt,
    right=6pt,
    top=5pt,
    bottom=5pt,
    title=\textbf{MGSM Prompt} \hfill \textit{Multilingual mathematical reasoning},
    fonttitle=\bfseries,
    coltitle=black,
    attach boxed title to top left={xshift=4pt,yshift=-2pt},
    boxed title style={
        colback=blue!10,
        colframe=blue!45!black,
        boxrule=0.4pt,
        arc=2pt
    }
]
\begin{Verbatim}[
    fontsize=\footnotesize,
    breaklines=true,
    breakanywhere=true,
    breaksymbolleft={},
    breaksymbolright={}
]
Answer the following math problem. The last line of your response should be of the following format: 'Answer: $NUMBER' (without quotes) where NUMBER is the final integer answer.

{input_data}
\end{Verbatim}
\end{tcolorbox}

\vspace{0.45em}

\begin{tcolorbox}[
    enhanced,
    width=0.96\textwidth,
    colback=green!2,
    colframe=green!45!black,
    boxrule=0.55pt,
    arc=2pt,
    left=6pt,
    right=6pt,
    top=5pt,
    bottom=5pt,
    title=\textbf{GPQA Diamond Prompt} \hfill \textit{Scientific knowledge reasoning},
    fonttitle=\bfseries,
    coltitle=black,
    attach boxed title to top left={xshift=4pt,yshift=-2pt},
    boxed title style={
        colback=green!10,
        colframe=green!45!black,
        boxrule=0.4pt,
        arc=2pt
    }
]
\begin{Verbatim}[
    fontsize=\footnotesize,
    breaklines=true,
    breakanywhere=true,
    breaksymbolleft={},
    breaksymbolright={}
]
Answer the following multiple choice question. The last line of your response should be of the following format: 'Answer: $LETTER' (without quotes) where LETTER is one of ABCD. Think step by step before answering.

{input_data}
\end{Verbatim}
\end{tcolorbox}

\vspace{0.45em}

\begin{tcolorbox}[
    enhanced,
    width=0.96\textwidth,
    colback=orange!2,
    colframe=orange!55!black,
    boxrule=0.55pt,
    arc=2pt,
    left=6pt,
    right=6pt,
    top=5pt,
    bottom=5pt,
    title=\textbf{HumanEval Prompt} \hfill \textit{Code generation},
    fonttitle=\bfseries,
    coltitle=black,
    attach boxed title to top left={xshift=4pt,yshift=-2pt},
    boxed title style={
        colback=orange!12,
        colframe=orange!55!black,
        boxrule=0.4pt,
        arc=2pt
    }
]
\begin{Verbatim}[
    fontsize=\footnotesize,
    breaklines=true,
    breakanywhere=true,
    breaksymbolleft={},
    breaksymbolright={}
]
You are an AI that only responds with python code, NOT ENGLISH. You will be given a function signature and its docstring. Write your full implementation (restate the function signature).

IMPORTANT: If the function uses type hints like List, Dict, etc., make sure to include "from typing import *" at the beginning.

Use a Python code block to write your response. For example:
```python
from typing import *

def hello():
    print("Hello world!")
{input_data}
\end{Verbatim}
\end{tcolorbox}

\end{figure*}

\end{document}